\documentclass[pdflatex,sn-nature]{sn-jnl}

\usepackage{graphicx}
\usepackage{multirow}
\usepackage{amsmath,amssymb}
\usepackage{booktabs}
\usepackage{xcolor}
\usepackage{url}


\begin{document}

\title[Evaluating Agentic Harness Systems for Autonomous Computational Pathology]{Evaluating Agentic Harness Systems for Autonomous Computational Pathology}

\author[1]{\fnm{Jie} \sur{Lin}}
\equalcont{These authors contributed equally to this work.}

\author[2]{\fnm{Zongyi} \sur{Chen}}
\equalcont{These authors contributed equally to this work.}

\author[3]{\fnm{Qiaoling} \sur{Zheng}}
\equalcont{These authors contributed equally to this work.}

\author[1]{\fnm{Liuyi} \sur{Wang}}

\author[1]{\fnm{Hengyi} \sur{Jiang}}

\author[2]{\fnm{Jiabao} \sur{Chen}}

\author[4]{\fnm{Xiang} \sur{Liu}}

\author*[3]{\fnm{Yinghong} \sur{Yang}}\email{yyh1555@163.com}

\author*[1,2]{\fnm{Liansheng} \sur{Wang}}\email{lswang@xmu.edu.cn}

\affil[1]{\orgdiv{Department of Computer Science at School of Informatics}, \orgname{Xiamen University}, \orgaddress{\city{Xiamen}, \country{China}}}

\affil[2]{\orgname{National Institute for Data Science in Health and Medicine, Xiamen University}, \orgaddress{\city{Xiamen}, \country{China}}}

\affil[3]{\orgdiv{Department of Pathology}, \orgname{Fujian Medical University Union Hospital}, \orgaddress{\city{Fuzhou}, \country{China}}}

\affil[4]{\orgdiv{Department of Artificial Intelligence at School of Informatics}, \orgname{Xiamen University}, \orgaddress{\city{Xiamen}, \country{China}}}

\abstract{Autonomous computational pathology (ACP) converts high-level pathology analysis goals into executable, traceable and clinically bounded workflows. Realizing this capability requires adapting general agentic harness systems to pathology-specific tasks, tools, evidence standards and clinical claim boundaries. We contribute ACP-Bench, a framework that adapts existing harness systems from computational pathology support toward ACP workflow capability. ACP-Bench evaluates 41 pathology workflow tasks, including 24 biomarker, 7 morphology and 10 prognosis tasks spanning 6 body-system groups and 9 endpoint families. The benchmark evaluates 9 models and 3 harness groups (Claude Code, Codex and Open Code), yielding 369 complete trajectories. ACP-Bench evaluates each trajectory across workflow execution, diagnostic performance and clinical-boundary alignment, combining expert-adjudicated process audits, diagnostic assessment and pathologist-validated safety review. Across evaluated systems, workflow initiation, task interpretation and diagnostic reporting were more mature than tool-bound execution, result binding and reflective workflow revision, and formal end-to-end completion remained rare. ACP-Bench provides a reusable standard for auditing whether agentic systems can operationalize pathology workflows before claims of reliable clinical autonomy.}


\maketitle

\section*{Introduction}

Agentic harness systems couple foundation models with planning, tool use, workflow memory, logs and feedback mechanisms. With their emergence, medical artificial intelligence is moving from static prediction toward agentic workflows that can interpret goals, execute tools and revise multi-step work. Generalist medical AI frames this transition as a move from fixed task models to flexible multimodal reasoning systems \cite{moor2023gmai,topol_high-performance_2019,rao_multimodal_2025,sellergren_medgemma_2025}. Computational pathology is a demanding test case for this shift, because pathology AI must connect image representation, clinical interpretation and human--AI workflow constraints \cite{litjens2017survey,huang_pathologistai_2025,faust_pharaoh_2025}. A single pathology goal can imply whole-slide image preprocessing, feature extraction, multiple-instance learning, task-target-specific evaluation, report generation and iterative correction.

We define autonomous computational pathology (ACP) as agentic systems that transform high-level pathology analysis goals into executable and clinically bounded computational workflows. This definition deliberately separates ACP from an autonomous pathologist. ACP extends pathology foundation-model evaluation from representation quality to workflow organization: beyond asking how well learned representations support biomarker, morphology or prognosis targets \cite{neidlinger2025pathologyfm,zhao_foundation_2025,truhn_large_2023}, it asks whether an agentic harness system can convert pathology knowledge into a working analysis chain.

The ingredients for ACP are becoming available. Whole-slide image frameworks make preprocessing, segmentation, feature extraction and multimodal integration more accessible \cite{lazyslide2026}. Biomedical data-science agents and multi-agent frameworks show that language models can plan, code and execute tool-aware analyses across specialized biomedical workflows \cite{wang2026biodsbench,bu2026biomedagent,li_mmedagent_2024,wang_spatialagent_2025}. Medical and imaging agents further show the appeal of modular tools, guideline grounding, intermediate visualization and human feedback in translational settings \cite{ferber2025oncologyagent,tissuelab2026,huang2025biomni,vaidya2025nova}. Scientific and molecular discovery agents extend the same logic to long-horizon experimentation and skill-structured tool orchestration \cite{lu2026aiscientist,molclaw2026,swanson_virtual_2025,boiko2023autonomous,wang2023scientific}.

These developments demonstrate the feasibility of agentic histopathology analysis, but they do not by themselves define the reusable procedural assets needed for clinically bounded workflow autonomy. ACP-Bench therefore treats pathology workflow skills as first-class evaluable objects, extending evaluation from task answers to decomposition, execution, evidence binding and revision.

These advances also expose a measurement gap. Output artifacts can look complete even when the underlying workflow remains weak. A final diagnostic report may be fluent without trace evidence that the system found and used the relevant results. A recovered prediction output may list labels without showing that the endpoint, metric and evidentiary scope matched the task. A self-critique or repair note may propose an update without producing a rerun artifact or measurable improvement. This gap is not unique to pathology. General agent benchmarks increasingly emphasize process-sensitive evaluation, deterministic checks, safety constraints and skill augmentation because output-only grading can miss how the result was produced \cite{claw_eval2026,skillsbench2026,liu2023agentbench,yao2023reactsynergizingreasoningacting,schick2023toolformer,mialon2023augmented,xie2024osworld,trivedi2024appworld,jimenez2024swebench}. Clinical evaluation proposals make a related point: static snapshots do not capture action, temporal state and operational consequence \cite{luo2026ces}.

Pathology sharpens this problem because the intermediate workflow helps define the clinical evidence boundary. Biomarker, morphology and prognosis tasks use different endpoints, reference labels and permissible claims. A retrospective whole-slide-image-to-label prediction can support a computational prediction claim, but it should not be treated as prospective pathologist adjudication or clinical diagnostic authority. A workflow that computes a plausible metric can still drift from the intended endpoint, omit supporting evidence or overstate safety. Thus ACP evaluation must inspect the chain from pathology goal to plan, action, diagnostic evidence, claim boundary and reflection.

ACP-Bench makes three contributions. First, it defines pathology-specific workflow skills that specify the core steps an agent must complete in computational pathology workflows. Second, it introduces three complementary evaluation dimensions for agentic pathology systems: automation ability, diagnostic result and clinical workflow. These dimensions separately measure whether a harness completes an executable workflow, whether recoverable whole-slide-image-to-label (WSI-to-label) outputs match reference labels, and whether task target, evidence support and clinical claim boundaries are preserved. Clinical workflow alignment was measured by CWAS-3, which scores unsupported completion, endpoint, label or metric mismatch, and missing evidence. Safety-boundary signals were measured by R6, a retrospective review layer covering endpoint drift, unsafe overclaim, conflict evidence, temporal or source misassignment and reflection safety. Structured pathologist validation then assessed whether selected reports preserved endpoint alignment, evidence sufficiency and safe claim-boundary language. Third, ACP-Bench converts these assessments into quantitative capability profiles, operational failure modes and clinical-boundary analyses rather than a single leaderboard score.

We developed ACP-Bench to evaluate whether agentic systems can convert pathology goals into executable and clinically bounded workflows. The benchmark spans 41 pathology workflow tasks, 9 agentic harness system groups and 369 complete workflow trajectories. Each trajectory was assessed by an expert-adjudicated rubric over planning, action, diagnosis and reflection. We then added qualitative failure analysis, WSI-to-label diagnostic-result metrics, CWAS-3 clinical workflow alignment, R6 safety-boundary review, pathologist validation, reflection validation and task-complexity analysis.

The result is a conservative benchmark finding. Evaluated harness systems showed emerging ACP capability, but not reliable end-to-end autonomy. The overall expert-adjudicated workflow score was 61.53\%, and only 10 of 369 trajectories reached the 75\% formal pass threshold. Planning and diagnostic reporting were stronger than tool-bound execution, result discovery and reflection. Diagnostic-result accuracy, clinical workflow alignment, safety-boundary review and pathologist validation measured different properties of the same trajectories, so these evidence layers could not be substituted for one another. ACP-Bench therefore provides a benchmark framework for evaluating whether agentic pathology systems can convert high-level goals into clinically bounded computational workflows.

\section*{Results}

In this section, we present the main ACP-Bench findings. We first define the benchmark task space and the trajectory-level evaluation axes. We then report workflow completion, recovered prediction performance and clinical claim-boundary preservation. Finally, we examine reflection rounds, long-horizon task demand and selected workflow examples to localize the observed gaps in ACP capability.

\subsection*{ACP-Bench evaluates complete workflow trajectories}

\begin{figure}[!p]
    \centering
    \includegraphics[width=\textwidth,height=0.78\textheight,keepaspectratio]{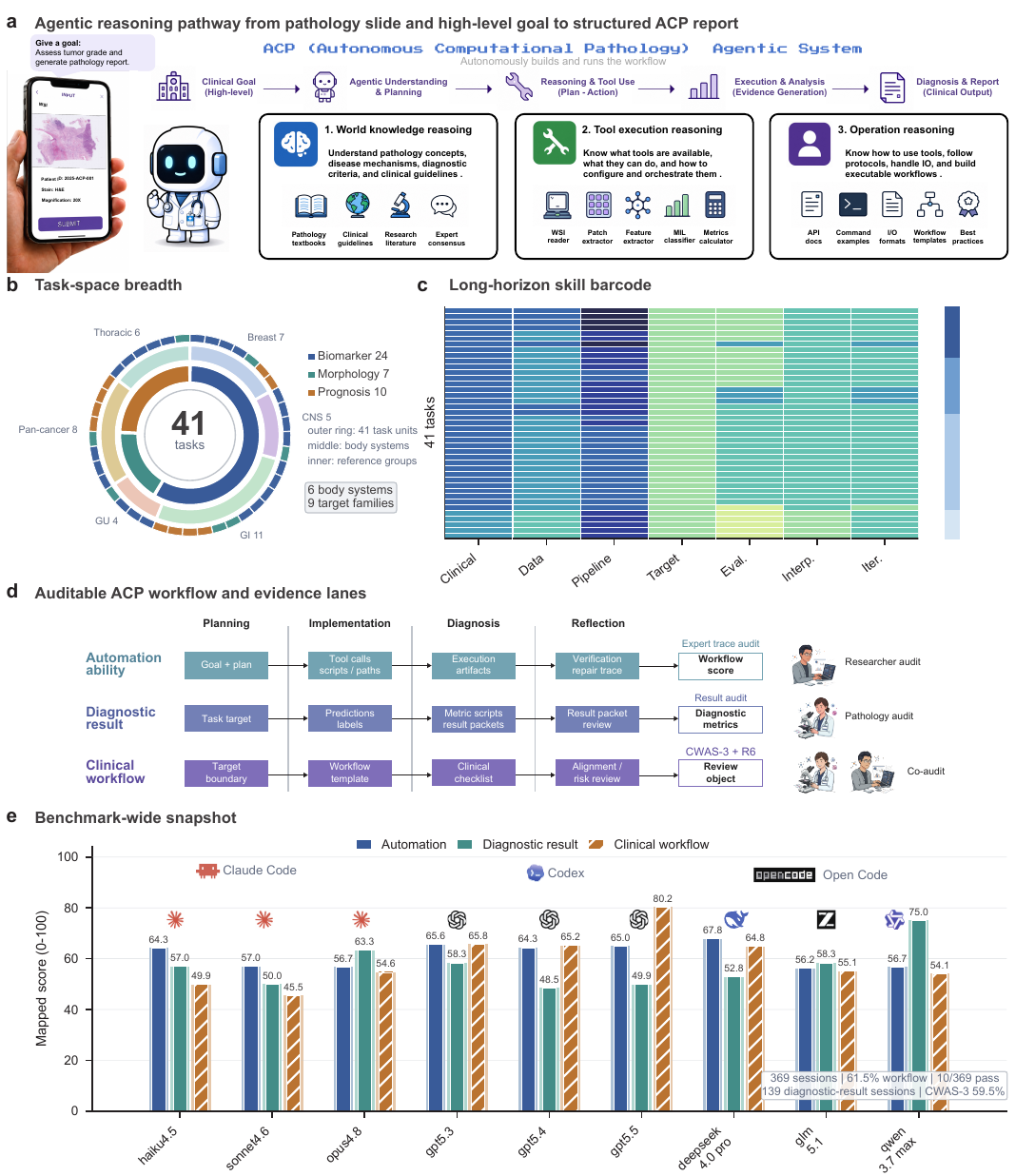}
    \caption{\textbf{ACP-Bench operationalizes autonomous computational pathology as an evaluable workflow.} \textbf{a,} Agentic reasoning pathway from pathology slide and high-level goal to structured ACP report. \textbf{b,} Task-space breadth across 41 pathology workflow tasks. \textbf{c,} Compact summary of long-horizon task demand, defined as the mapped workflow requirements needed by a task. \textbf{d,} Three evidence lanes used by the benchmark: automation ability, diagnostic result and clinical workflow. \textbf{e,} Benchmark-wide snapshot across 9 harness/model groups, with scores mapped to 0--100. Group displays are descriptive harness-level evidence profiles. Scale badges report the 369 trajectories, 10 formal workflow passes, 139 trajectories with diagnostic-result metrics and CWAS-3 mean of 59.5\%.}
    \label{fig:overview}
\end{figure}

ACP-Bench tested whether a high-level pathology goal became an executable and clinically bounded computational workflow. The benchmark contained 41 pathology workflow tasks, 9 harness/model groups across 3 harness families and 369 complete workflow trajectories (Fig.~\ref{fig:overview}). Each task was described by task metadata, including organ, body system, target family, label-reference type, output structure, metric family, defined as the expected evaluation metric class, and clinical claim boundary.

The evaluation unit was the complete workflow trajectory. ACP-Bench represented each trajectory through planning, action, diagnosis and reflection stages. The same trajectory was then evaluated along three axes: whether the workflow was executable and inspectable, whether recovered WSI-to-label predictions were correct, and whether the output preserved the intended clinical claim boundary.

Automation ability was measured for all 369 trajectories with a rubric over planning, action, diagnosis and reflection. Diagnostic-result performance was measured for the subset with paired reference and predicted labels, comprising 139 trajectories and 1,743 normalized prediction rows, where each row paired a harmonized reference label with a predicted label. Clinical-workflow alignment was measured for all 369 trajectories with CWAS-3, a three-component clinical workflow alignment score, and R6, a retrospective safety-boundary review index. These axes used different denominators and scoring rules and were reported as separate benchmark measurements.

\clearpage

\subsection*{Task metadata define the benchmark task space}

\begin{figure}[!p]
    \centering
    \includegraphics[width=\textwidth,height=0.78\textheight,keepaspectratio]{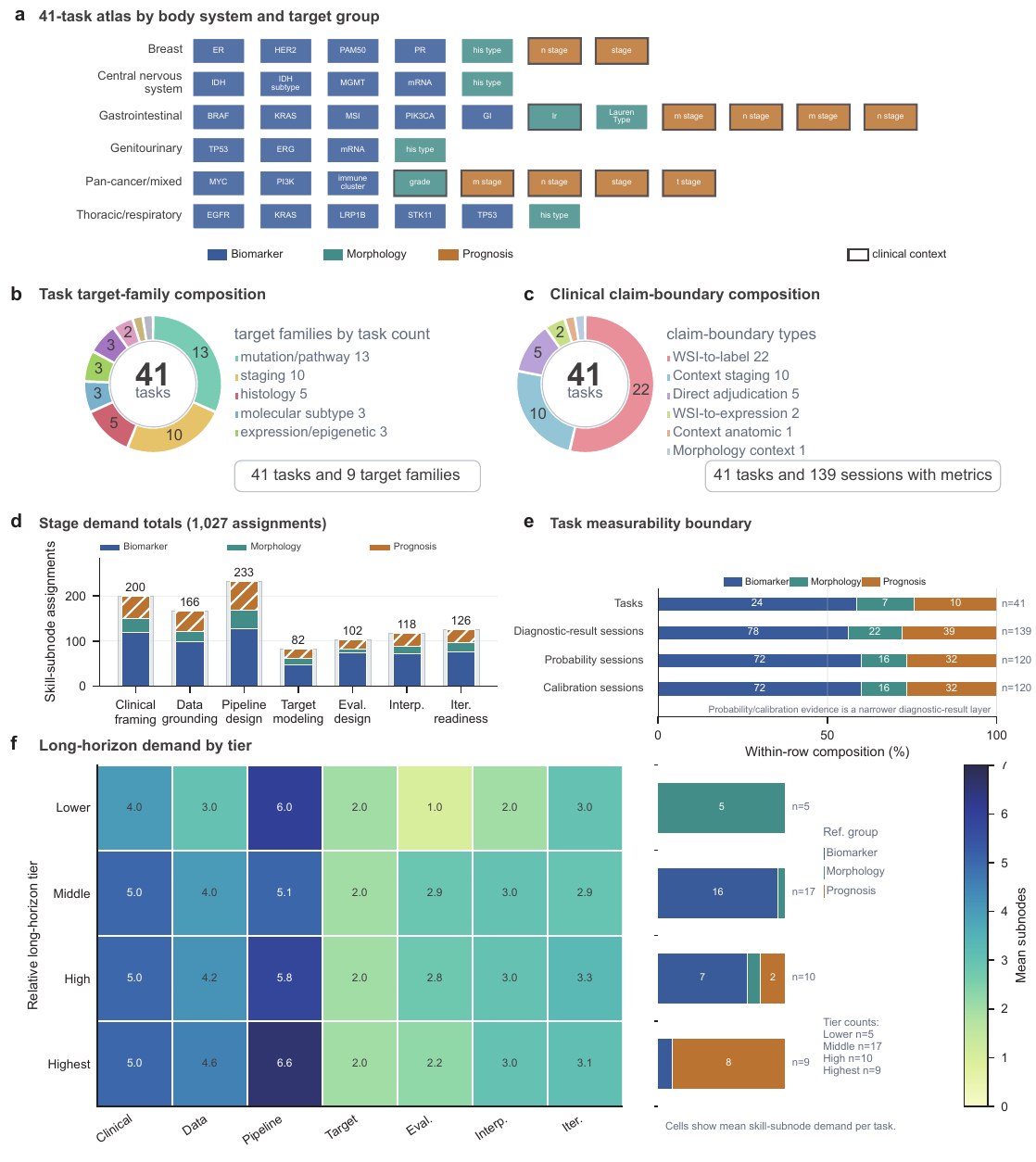}
    \caption{\textbf{ACP-Bench tasks form a workflow-demand summary.} \textbf{a,} Task atlas by body system, with tile color denoting reference group and outlined tiles denoting clinical-context requirements. Reference groups are biomarker, morphology and prognosis. \textbf{b,c,} Task target-family and clinical claim-boundary composition across 41 pathology workflow tasks; diagnostic-result badges in these panels refer to trajectory-level metric availability. \textbf{d,} Workflow-requirement subnode assignment counts by workflow stage and reference group. Subnodes are audit-defined workflow requirements and are described as such throughout the manuscript. \textbf{e,} Metric-availability boundary linking 41 tasks to narrower diagnostic-result evidence. The 139 count denotes diagnostic-result trajectories, the 1,743 count denotes normalized prediction rows, the 120 count denotes trajectories with probability outputs and the 19 count denotes probability-metric exclusions. \textbf{f,} Long-horizon tier by stage-demand heatmap with reference-group composition by tier. Counts describe task-demand assignments.}
    \label{fig:task_demand}
\end{figure}

The task metadata defined the pathology task space used for ACP evaluation. ACP-Bench covered 6 body systems and 9 task target families, with 24 biomarker tasks, 7 morphology tasks and 10 prognosis tasks (Fig.~\ref{fig:task_demand}a,b). The task framing followed computational pathology benchmark practice, where biomarker, morphology and prognosis targets are linked to slide-level labels and downstream metric families \cite{neidlinger2025pathologyfm,bareja2025evaluating}. Metric family denotes the expected evaluation metric class for a task, such as AUROC/AUC, AUPRC, F1, balanced accuracy or accuracy. The 41 tasks were selected when they could be represented as pathology goals with an organ or pan-cancer context, a reference label source, an output structure and a metric family.

Task target families were assigned by the reference label and recoverable output. Biomarker and prognosis tasks were treated as WSI-to-label prediction tasks because their recoverable outputs were slide-level molecular, biomarker or outcome labels. A task was treated as direct diagnostic adjudication only when a pathologist-adjudicated diagnostic label was available as the reference. Morphology tasks were kept as a separate target family because they asked about tissue-pattern or phenotype labels.

We also mapped each task to seven audit-defined workflow requirement categories: clinical framing, data-source grounding, pipeline design, target modeling, evaluation design, interpretation and iteration readiness. Data-source grounding denotes the required slide, label, split and metadata inputs. The other categories captured the clinical context, executable analysis steps, endpoint-specific modeling, metric compatibility, report requirements and rerun readiness. The resulting task demand was the set of required workflow categories and subnodes assigned to each task. Mapping the 41 tasks to these categories produced 1,027 requirement assignments. Pipeline design carried the largest total demand (233 assignments), followed by clinical framing (200), data-source grounding (166), iteration readiness (126), interpretation (118), evaluation design (102) and target modeling (82) (Fig.~\ref{fig:task_demand}d).

We next quantified metric availability and task complexity. Diagnostic-result metrics covered 139 trajectories and 1,743 normalized prediction rows. Probability metrics covered 120 trajectories with probability outputs, after excluding 19 trajectories whose probability columns could not be conservatively mapped. Relative long-horizon tiers captured longer workflow chains and showed different stage-demand profiles and reference-group composition (Fig.~\ref{fig:task_demand}e,f). Reference groups denoted the benchmark target classes: biomarker, morphology and prognosis.

Together, these components linked pathology goals to the task metadata, workflow requirements and measurement denominators used throughout ACP-Bench.

\clearpage

\subsection*{Harness systems rarely completed full workflows}

\begin{figure}[!p]
    \centering
    \includegraphics[width=\textwidth,height=0.82\textheight,keepaspectratio]{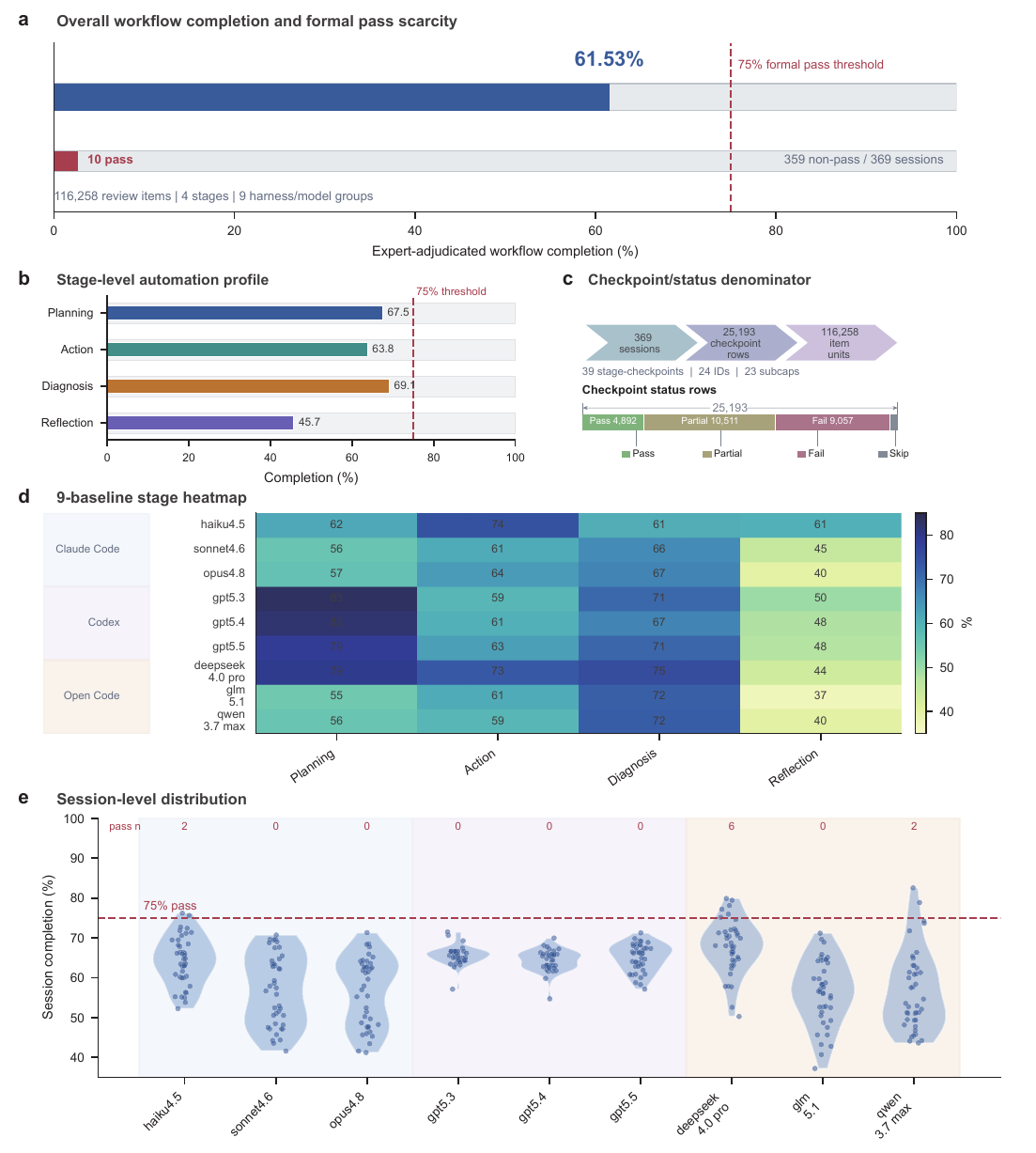}
    \caption{\textbf{Evaluated harness systems showed partial but unreliable workflow completion.} \textbf{a,} Overall expert-adjudicated workflow score and formal pass count across 369 trajectories. \textbf{b,} Stage-level automation profile for planning, action, diagnosis and reflection. \textbf{c,} Checkpoint/status denominator view, including item coverage and status classes. \textbf{d,} Nine-group stage heatmap organized by harness family. \textbf{e,} Trajectory-level completion distributions with the 75\% formal pass threshold and pass counts. Baseline panels are descriptive harness-level summaries. Scores report workflow assessment.}
    \label{fig:expert_audit}
\end{figure}

We next evaluated whether harness systems completed the workflow steps required by each task. The assessment used an expert-adjudicated rubric over planning, action, diagnosis and reflection for all 369 trajectories. Evaluated harness systems showed partial ACP capability but rarely met the formal workflow-completion threshold. The overall rubric-based workflow score was 61.53\%, and 10 of 369 trajectories reached the 75\% formal pass threshold (Fig.~\ref{fig:expert_audit}a).

Stage-level scores localized the weakness. Diagnosis scored 69.13\%, planning scored 67.48\%, action scored 63.77\% and reflection scored 45.73\% (Fig.~\ref{fig:expert_audit}b). The strongest stages were closest to task interpretation and diagnostic reporting, and the weakest stage was iterative reflection. Baseline-stage profiles varied across harness families, and no group was uniformly strong across all four stages (Fig.~\ref{fig:expert_audit}d,e). Many systems could start workflows and produce reports, but fewer trajectories executed tools, located outputs, verified results and completed reflection rounds with comparable metric evidence.

The assessment combined reproducible output checks with reviewer judgments constrained to evidence from each trajectory. Across 116,258 review items, 50,277 were deterministic checks over workflow outputs, scripts, metrics or logs, and 65,981 were semantic checkpoint judgments. To prevent report-only scoring, a final narrative report could not satisfy a checkpoint unless the required artifact, metric or trace evidence from the same workflow run was present. At pass thresholds of 70\%, 75\% and 80\%, 35, 10 and 1 of 369 trajectories passed, respectively. All three thresholds showed that complete workflow execution was uncommon.

Semantic checkpoint judgments were reproducible enough to support aggregate interpretation but retained boundary ambiguity. A pilot calibration review covered 9 trajectories and 2,738 item judgments, with 83.27\% exact agreement against the initial adjudication set before rule refinement. A pre-specified independent reliability review then sampled 27 trajectories across score strata and harness/model groups. Three reviewers independently judged 4,320 semantic checkpoint items using the same status policy. All-three exact agreement was 59.7\% overall, and pairwise weighted Cohen kappa ranged from 0.560 to 0.672. The primary three-rater interval Krippendorff alpha was 0.658, with the highest residual ambiguity in action, diagnosis and reflection checkpoint boundaries.

\clearpage

\subsection*{Execution and repair gaps explain low workflow completion}

\begin{figure}[!p]
    \centering
    \includegraphics[width=\textwidth,height=0.82\textheight,keepaspectratio]{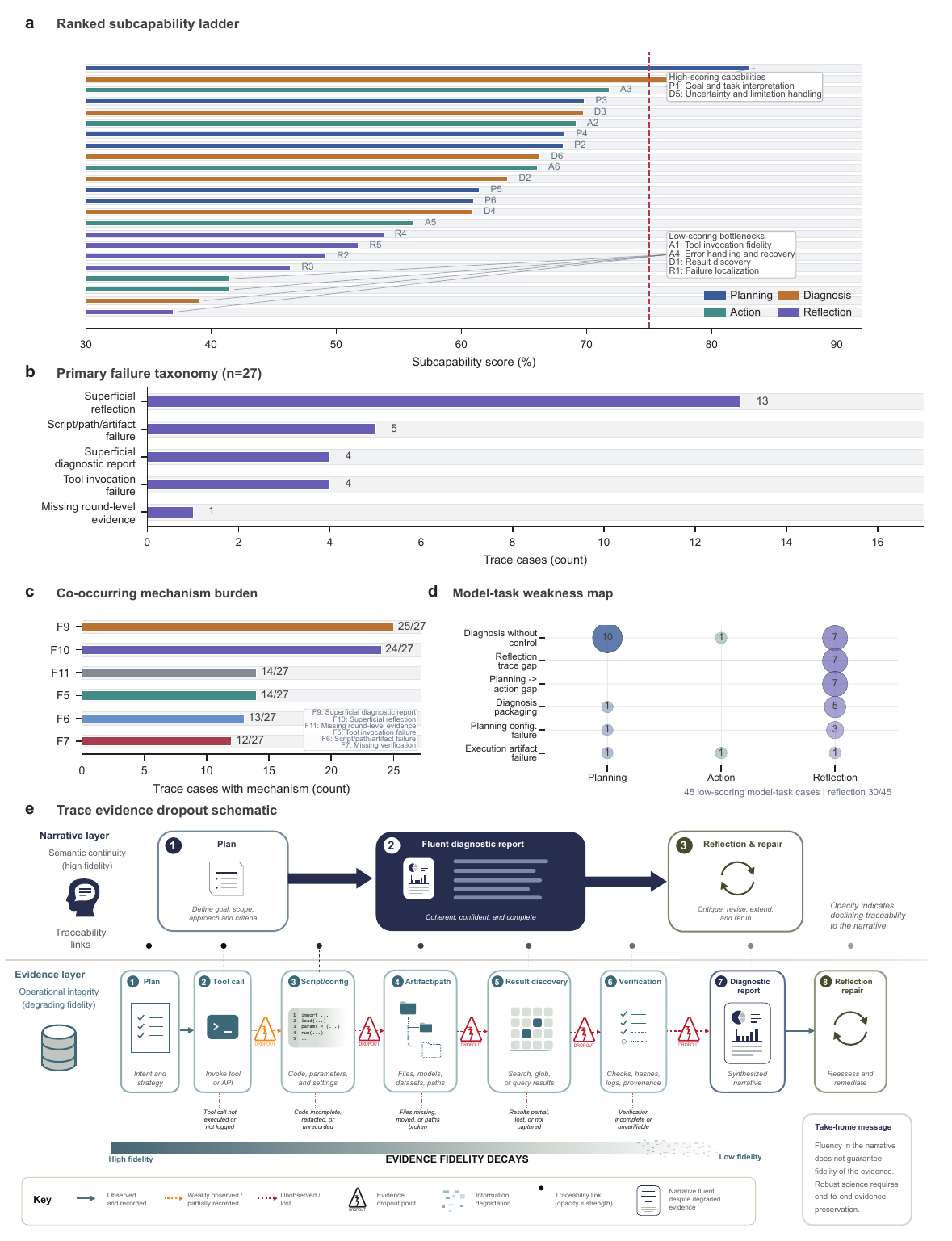}
    \caption{\textbf{Failure analysis identifies workflow-conversion bottlenecks.} \textbf{a,} Ranked subcapability scores, with stage color indicating planning, action, diagnosis or reflection. Subcapability codes denote workflow-stage groups: P for planning, A for action, D for diagnosis/reporting and R for reflection. Thus D1 and D5 are diagnosis-stage subcapabilities, whereas reflection-related subcapabilities use the R prefix. \textbf{b,} Primary failure labels in 27 sampled process cases. \textbf{c,} Primary plus secondary failure mechanisms, showing that failures often co-occurred. \textbf{d,} Model-task weakness summary from 45 low-scoring model-task cases, separate from the 27-case qualitative taxonomy. \textbf{e,} Evidence-dropout schematic linking weak tool invocation, output binding, verification and round evidence to diagnostic reporting without trace control. The 27 cases explained mechanisms behind aggregate scores; the 45-case weakness summary pinpointed low-score patterns.}
    \label{fig:failure}
\end{figure}

Subcapability scores showed that models understood tasks more often than they completed executable workflows. These scores grouped checkpoint items within workflow stages and averaged adjudicated scores over eligible trajectories. Goal and task interpretation reached 83.04\%, and uncertainty or limitation handling reached 80.34\% (Fig.~\ref{fig:failure}a). Lower-scoring subcapabilities were tool invocation fidelity (41.49\%), error handling or recovery (41.49\%), result discovery (39.06\%) and reflection-stage failure localization (36.95\%). This profile separated task understanding and report wording from tool-bound execution and repair control.

Qualitative process review identified the mechanisms behind these score patterns. The review sampled 27 cases, selecting one high-score, one middle-score and one low-score trajectory from each of the 9 harness/model groups. Superficial reflection, defined as reflection text without a workflow-evidenced failure or documented next-round change, was the most common primary failure label, appearing in 13 of 27 cases (Fig.~\ref{fig:failure}b). When primary and secondary labels were combined, superficial diagnostic report, defined as diagnostic-sounding text without sufficient supporting workflow or metric evidence, appeared in 25 of 27 cases and superficial reflection appeared in 24 of 27 cases (Fig.~\ref{fig:failure}c). Missing round-level evidence appeared in 14 of 27 cases, tool invocation failure in 14 of 27, script or output failure in 13 of 27 and missing verification in 12 of 27. A separate 45-row low-scoring model-task summary identified recurrent weak stages and mechanisms (Fig.~\ref{fig:failure}d). Many low-scoring trajectories still produced diagnostic-sounding reports despite missing tool execution, output discovery or verification evidence (Fig.~\ref{fig:failure}e).

\clearpage

\subsection*{Workflow, prediction and clinical-boundary scores diverged}

Workflow completion alone left clinical alignment unresolved. This analysis used three denominators. Automation ability, CWAS-3 and R6 were scored across all 369 trajectories. Diagnostic-result metrics used the subset with paired reference and predicted labels, comprising 139 trajectories and 1,743 normalized prediction rows. Probability metrics used 120 trajectories with probability outputs after the probability-column exclusions described above.

\begin{figure}[!p]
    \centering
    \includegraphics[width=\textwidth,height=0.76\textheight,keepaspectratio]{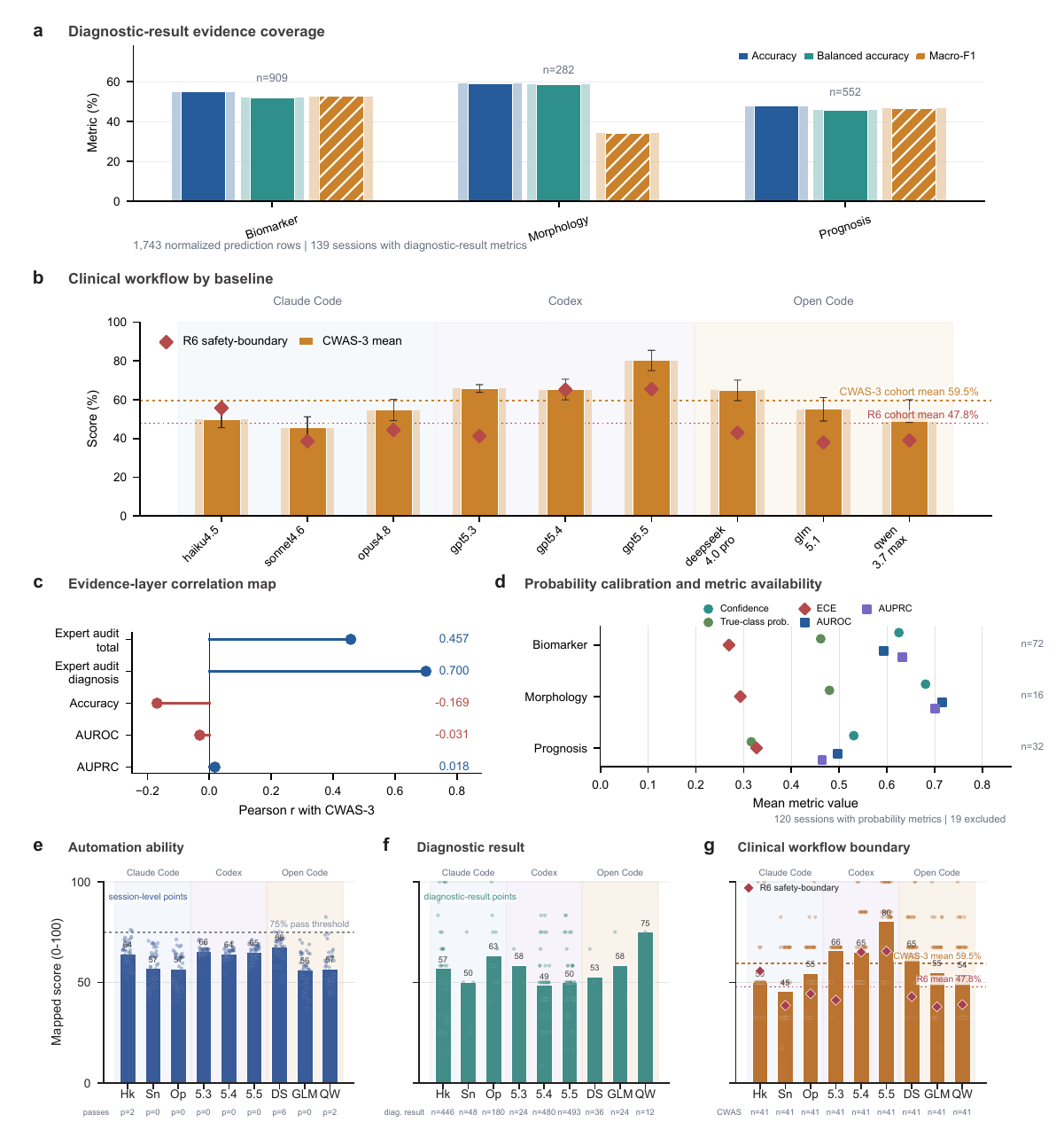}
    \caption{\textbf{Diagnostic result and clinical workflow alignment are complementary evidence layers.} \textbf{a,} Accuracy, balanced accuracy and macro-F1 by reference group for 1,743 normalized prediction rows from 139 diagnostic-result trajectories. \textbf{b,} CWAS-3 and R6 summaries by harness/model group across 369 trajectories. \textbf{c,} Correlation map relating CWAS-3 to expert assessment and diagnostic-result metrics. \textbf{d,} Probability calibration and metric availability by reference group among 120 trajectories with probability outputs, with 19 normalized trajectories excluded from probability metrics because probability columns lacked conservative task-label mappings. \textbf{e,} Automation ability by harness/model group across 369 trajectories. \textbf{f,} Diagnostic-result accuracy by harness/model group using normalized prediction-row denominators. \textbf{g,} Clinical workflow alignment by harness/model group across 369 trajectories, with R6 shown as a safety-boundary overlay; short x-axis codes follow the group order in panel b. Panels e--g are descriptive harness-level evidence profiles. CWAS-3 summarizes workflow alignment and R6 summarizes safety-boundary review signals.}
    \label{fig:clinical}
\end{figure}

Normalized diagnostic-result metrics evaluated WSI-to-label prediction rows when paired reference and predicted labels were recoverable. Overall discrete-label accuracy was 0.532, balanced accuracy was 0.516 and macro-F1 was 0.352 (Fig.~\ref{fig:clinical}a). These values summarize recovered prediction outputs in the normalized subset; the pathologist review below assessed report-level claim-boundary behavior.

Clinical-boundary scoring showed a separate weakness. CWAS-3 combined unsupported completion, endpoint, label or metric mismatch, and missing evidence into a 0--100 clinical workflow alignment score. The mean CWAS-3 score across all 369 trajectories was 59.5\%. R6 summarized retrospective safety-boundary review signals across endpoint drift, unsafe overclaim, conflict evidence, temporal or source misassignment and reflection safety. The mean R6 score was 47.8\% (Fig.~\ref{fig:clinical}b). R6 flags served as review-queue signals for possible clinical-boundary deviations.

The correlation pattern also showed that the axes captured distinct trajectory properties. CWAS-3 correlated with expert-assessment total score (Pearson $r=0.457$) and more strongly with expert-assessment diagnosis ($r=0.700$). It showed negative or near-zero correlations with diagnostic-result accuracy ($r=-0.169$), AUROC ($r=-0.031$) and AUPRC ($r=0.018$) in trajectories where those metrics were available (Fig.~\ref{fig:clinical}c). Baseline-level displays changed when scores were separated into automation, diagnostic-result and clinical-workflow panels (Fig.~\ref{fig:clinical}d--g). ACP performance therefore required separate reporting of workflow execution, prediction quality and clinical-boundary alignment.

We next clinically validated report-level claim-boundary behavior through structured pathologist review of 90 selected reports enriched for potential clinical-boundary risk (Table~\ref{tab:pathologist_validation}). The review domains assessed whether the report retained the intended endpoint, kept the allowed claim scope, had enough evidence for its wording, avoided overclaim and would be safe for downstream use. Endpoint alignment was usually retained: 85 of 90 reviewed outputs were aligned with the intended endpoint, 4 were partially aligned and 1 was misaligned. The more frequent failure was that reports often made statements stronger than the available supporting evidence. Claim boundaries were fully preserved in 5 of 90 reviewed outputs, partially preserved in 74 and not preserved in 11. Evidence was judged insufficient for the report wording in 74 of 90 outputs, and unsafe overclaim was possible in 60 and present in 26. Clinical risk was moderate in 42 outputs and major or unacceptable in 35; recommended use was ``requires rework before review'' for 57 and ``do not use'' for 17.

These pathologist-validated findings supported the clinical-workflow lane. Many outputs preserved the nominal endpoint while failing to preserve evidence sufficiency, limitations or safe claim-boundary language. ACP failures therefore included incomplete workflows, incorrect labels and clinically meaningful boundary failures in how computational evidence was converted into report text.

\begin{table*}[t]
\centering
\caption{\textbf{Pathologist clinical validation of selected case-report safety and claim-boundary preservation.} Counts summarize 90 risk-enriched selected reviewed case-report packets and describe the reviewed set. The validation assessed clinical-boundary and report-safety behavior within selected outputs.}
\label{tab:pathologist_validation}
\small
\begin{tabular}{p{0.31\textwidth}p{0.24\textwidth}p{0.35\textwidth}}
\toprule
Review domain & Result & Interpretation \\
\midrule
Endpoint alignment & Aligned 85/90; partially aligned 4/90; misaligned 1/90 & Most reviewed outputs preserved the nominal target, so clinical-boundary failures were usually not simple endpoint substitutions. \\
\midrule
Claim-boundary preservation & Preserved 5/90; partially preserved 74/90; not preserved 11/90 & Reports commonly retained only partial control of the intended evidence boundary. \\
\midrule
Evidence sufficiency & Insufficient 74/90; partially sufficient 16/90 & The available workflow, prediction or metric evidence often did not support the report wording. \\
\midrule
Unsafe overclaim & No 4/90; possible 60/90; yes 26/90 & Potential or explicit overclaim was common in the reviewed outputs. \\
\midrule
Endpoint or normalization drift & No 85/90; possible 2/90; yes 3/90 & Drift was less frequent than evidence insufficiency or unsafe report wording. \\
\midrule
Conflicting evidence & No 43/90; possible 16/90; yes 31/90 & Nearly half of reviewed outputs contained possible or explicit conflicting evidence. \\
\midrule
Limitation handling & Partially adequate 70/90; inadequate 20/90 & Limitation statements were often incomplete even when some limitation language was present. \\
\midrule
Clinical risk level & Minor 13/90; moderate 42/90; major or unacceptable 35/90 & Most reviewed outputs required substantial caution before downstream use. \\
\midrule
Recommended use & Research workflow audit only 10/90; clinician reference only with review 6/90; requires rework before review 57/90; do not use 17/90 & Most reviewed outputs were not suitable as clinician-facing material without rework or exclusion. \\
\bottomrule
\end{tabular}
\end{table*}

\clearpage

\subsection*{Reflection and long-horizon analyses showed limited measurable improvement}

\begin{figure}[!p]
    \centering
    \includegraphics[width=\textwidth,height=0.80\textheight,keepaspectratio]{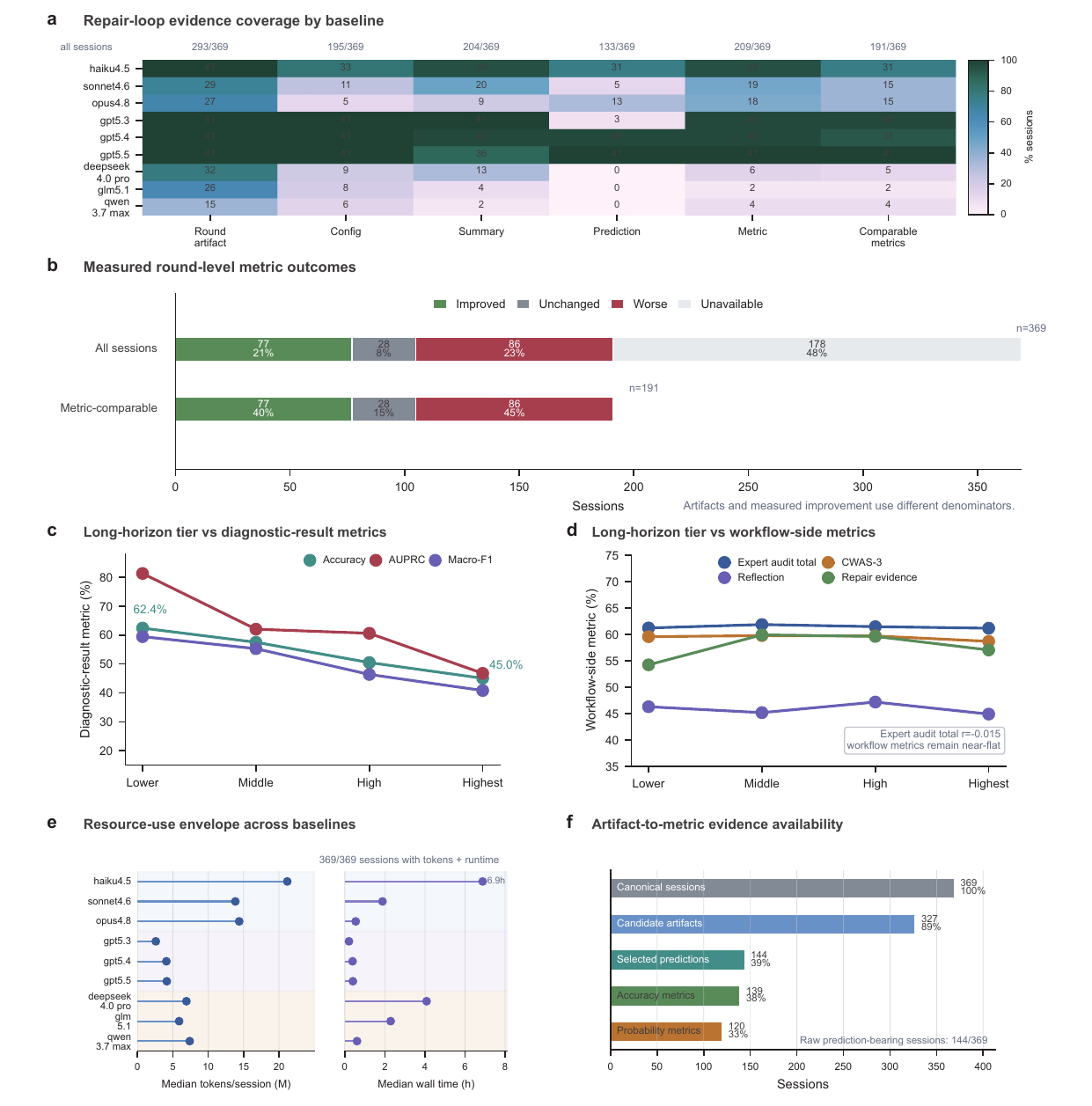}
    \caption{\textbf{Reflection metrics and long-horizon task demand show limited measurable improvement.} \textbf{a,} Reflection evidence coverage across all 369 trajectories. \textbf{b,} Measured round-level metric outcomes, shown for all trajectories and for the metric-comparable subset. \textbf{c,} Long-horizon tier versus diagnostic-result metrics. \textbf{d,} Long-horizon tier versus workflow-side metrics, with correlation summary. \textbf{e,} Computation-use envelope from complete 369-trajectory token and wall-time records. \textbf{f,} Evidence-availability funnel from candidate outputs to selected predictions, normalized metrics and probability metrics. Reflection outputs, measured improvement and long-horizon associations use the denominators shown in each panel and report retrospective associations rather than causal effects.}
    \label{fig:repair_longhorizon}
\end{figure}

Reflection analysis focused on whether later workflow rounds produced comparable metric evidence. Across 369 trajectories, 293 had at least one round output and 191 had at least two parsed round metrics (Fig.~\ref{fig:repair_longhorizon}a). Rubric-level reflection support was strong for 174 trajectories, partial for 74 and weak for 121. Among all trajectories, 77 improved, 28 were unchanged, 86 worsened and 178 lacked safely parsed comparable round-level metrics (Fig.~\ref{fig:repair_longhorizon}b). Reflection attempts were common, while metric-supported improvement was narrower and mixed. The reflection layer therefore exposed a gap between producing another round of work and producing a measurable improvement.

Higher long-horizon demand was associated with lower diagnostic-result performance. Long-horizon demand was a rule-based task-complexity score for within-benchmark stratification. Mean accuracy declined from 0.624 in the lower relative tier to 0.450 in the highest tier (Fig.~\ref{fig:repair_longhorizon}c). AUPRC and macro-F1 also declined in the highest tier. By contrast, expert-assessment total score remained approximately 61--62\% across tiers, and CWAS-3 stayed near 59--60\% (Fig.~\ref{fig:repair_longhorizon}d). The long-horizon score had near-zero correlation with expert-assessment total ($r=-0.015$), and negative correlations with accuracy ($r=-0.282$) and AUPRC ($r=-0.429$). Computation-use and evidence-availability analyses showed complete token and wall-time records for 369 trajectories and a narrowing from candidate outputs to selected, normalized and prediction outputs with probability columns (Fig.~\ref{fig:repair_longhorizon}e,f).

\clearpage

\subsection*{Selected trace boards illustrate individual failure modes}

We used selected qualitative trace boards to show how aggregate failures appeared inside individual workflows. The boards combined roadmap evidence states, selected casebook examples, automation outputs, diagnostic rows, report excerpts and task-matched attention overlays. Roadmap evidence states were visual markers of whether selected workflow nodes had supporting evidence. Casebook examples were selected trajectories used for qualitative illustration, and attention overlays were model-attention visualizations rather than pathologist-validated tissue annotations.

\begin{figure}[!p]
    \centering
    \includegraphics[width=\textwidth,height=0.86\textheight,keepaspectratio]{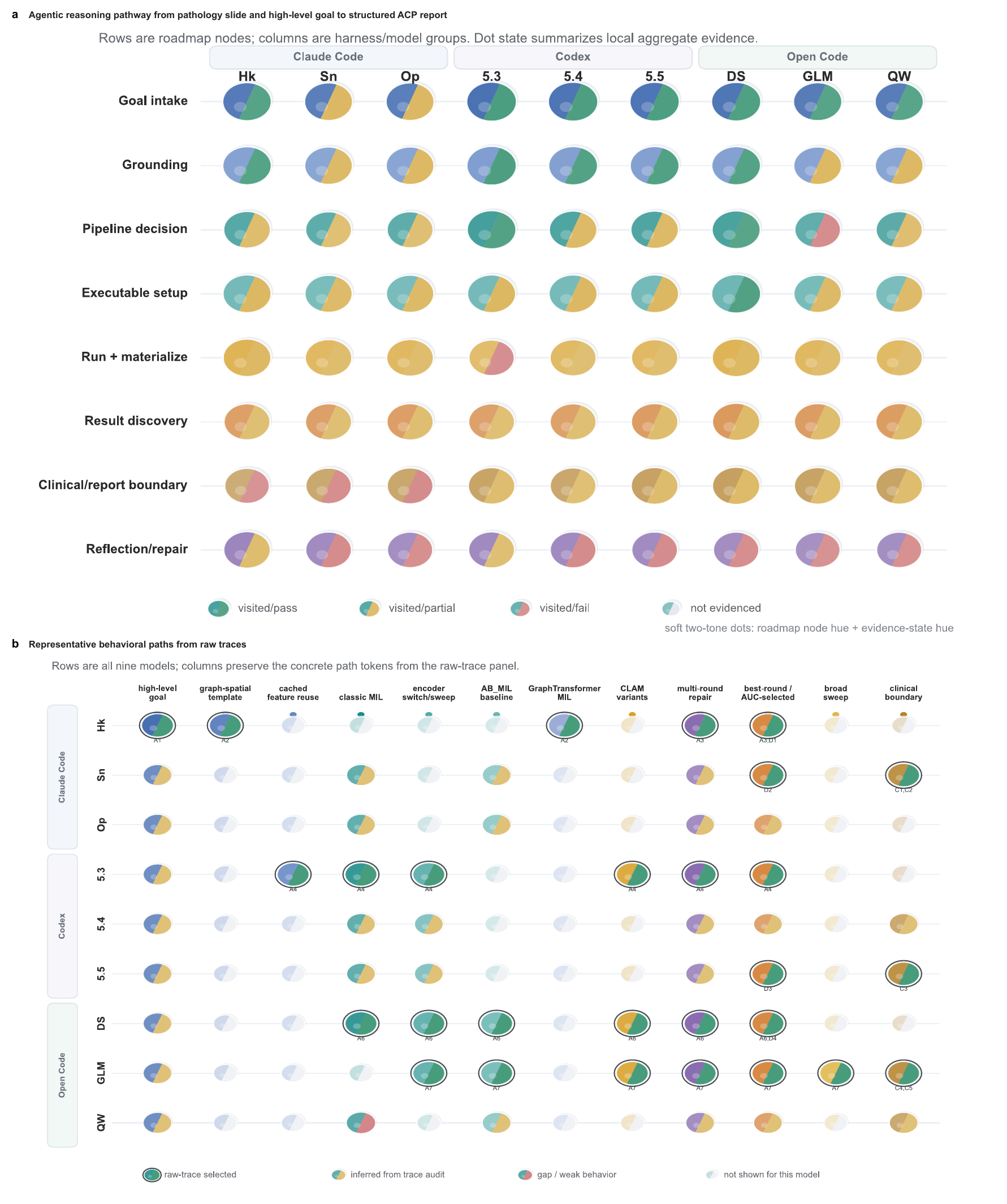}
    \caption{\textbf{Roadmap footprints and behavior paths provide qualitative evidence-state context.} The roadmap-footprint matrix summarizes assessment-derived evidence states for ACP workflow nodes across evaluated harness/model groups. The behavior-path layer adds selected casebook and directly observed evidence. Dot states mark evidence states. Soft or inferred dots indicate behavior supported by workflow-assessment summaries, selection across all harness/model groups or selected casebook evidence. The figure provides qualitative context for selected workflow behaviors.}
    \label{fig:roadmap_qual}
\end{figure}

\begin{figure}[!p]
    \centering
    \includegraphics[width=\textwidth,height=0.88\textheight,keepaspectratio]{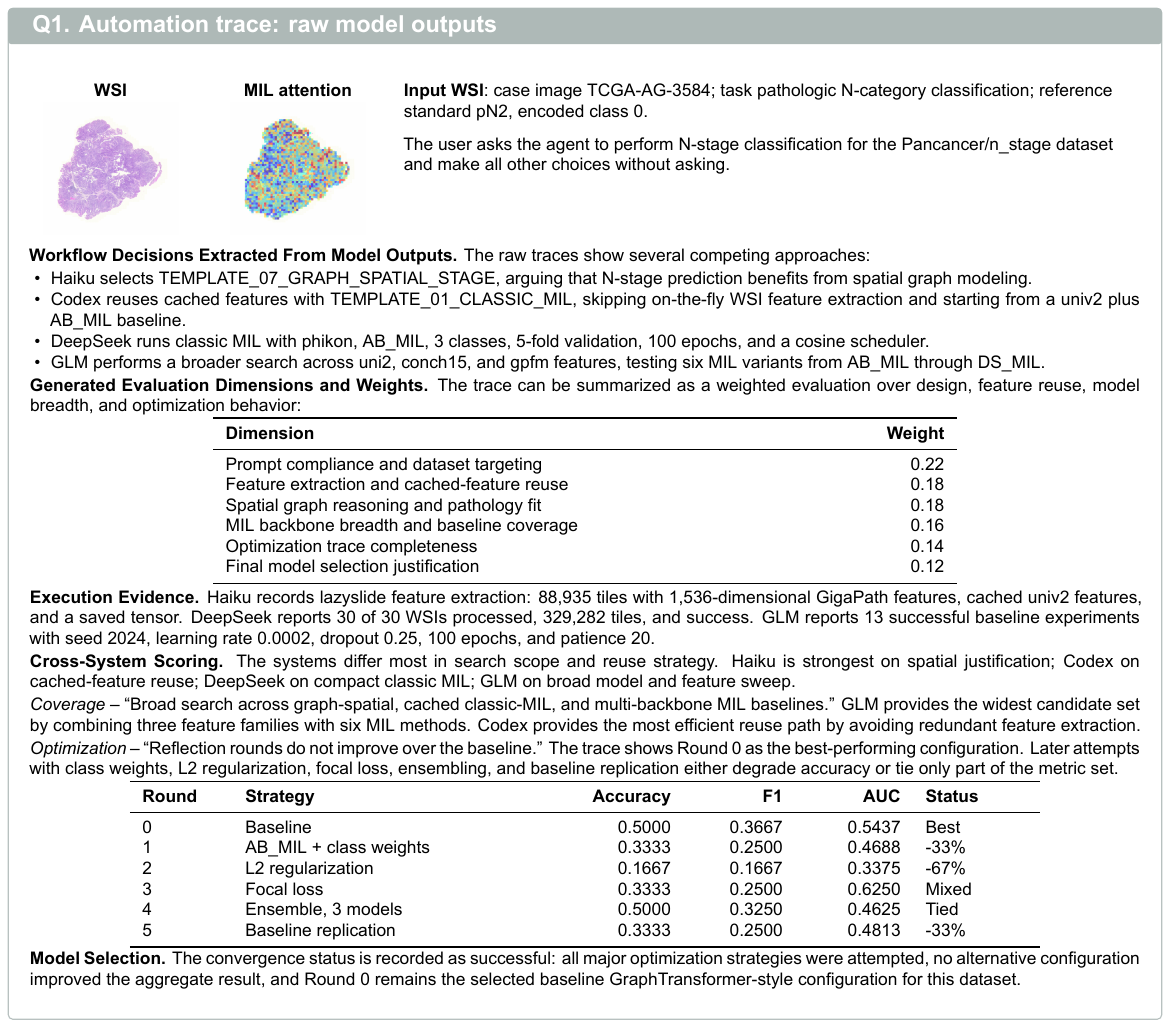}
    \caption{\textbf{Automation outputs show selected process evidence behind workflow scores.} The board combines selected workflow examples, including automation outputs, WSI thumbnails, multiple-instance-learning (MIL) attention context, reflection attempts and model-selection excerpts. It illustrates automation and reflection behavior in selected trajectories.}
    \label{fig:automation_qual}
\end{figure}

\begin{figure}[!p]
    \centering
    \includegraphics[width=\textwidth,height=0.88\textheight,keepaspectratio]{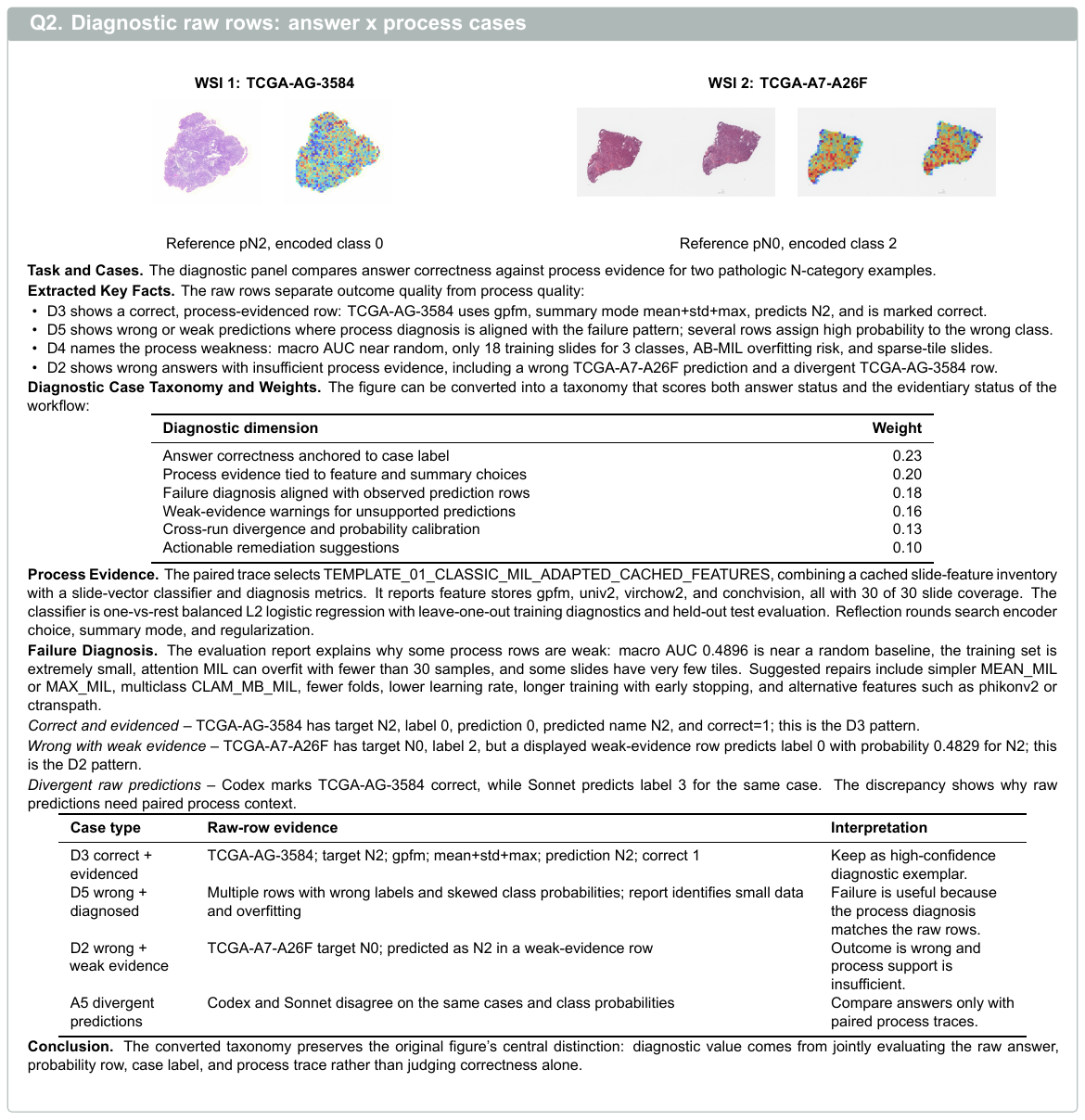}
    \caption{\textbf{Diagnostic rows show selected case-level evidence behind the failure taxonomy.} The board displays selected WSI cases from the qualitative casebook, reference standards, task-matched multiple-instance-learning (MIL) attention context and divergent predictions. It supports qualitative interpretation of Fig.~\ref{fig:failure} and Fig.~\ref{fig:clinical}.}
    \label{fig:diagnostic_qual}
\end{figure}

The roadmap and behavior-path board showed that evidence states varied across harness groups and workflow nodes (Fig.~\ref{fig:roadmap_qual}). The automation and diagnostic boards then mapped this variation to specific workflow steps: selected cases contained partial automation outputs, inconsistent model-selection text, divergent diagnostic rows and weak evidence binding (Figs.~\ref{fig:automation_qual}, \ref{fig:diagnostic_qual}). These examples made the aggregate failure labels interpretable as concrete workflow behaviors, especially result-discovery gaps, output-verification gaps and weak-evidence answer patterns.

\begin{figure}[!p]
    \centering
    \includegraphics[width=\textwidth,height=0.88\textheight,keepaspectratio]{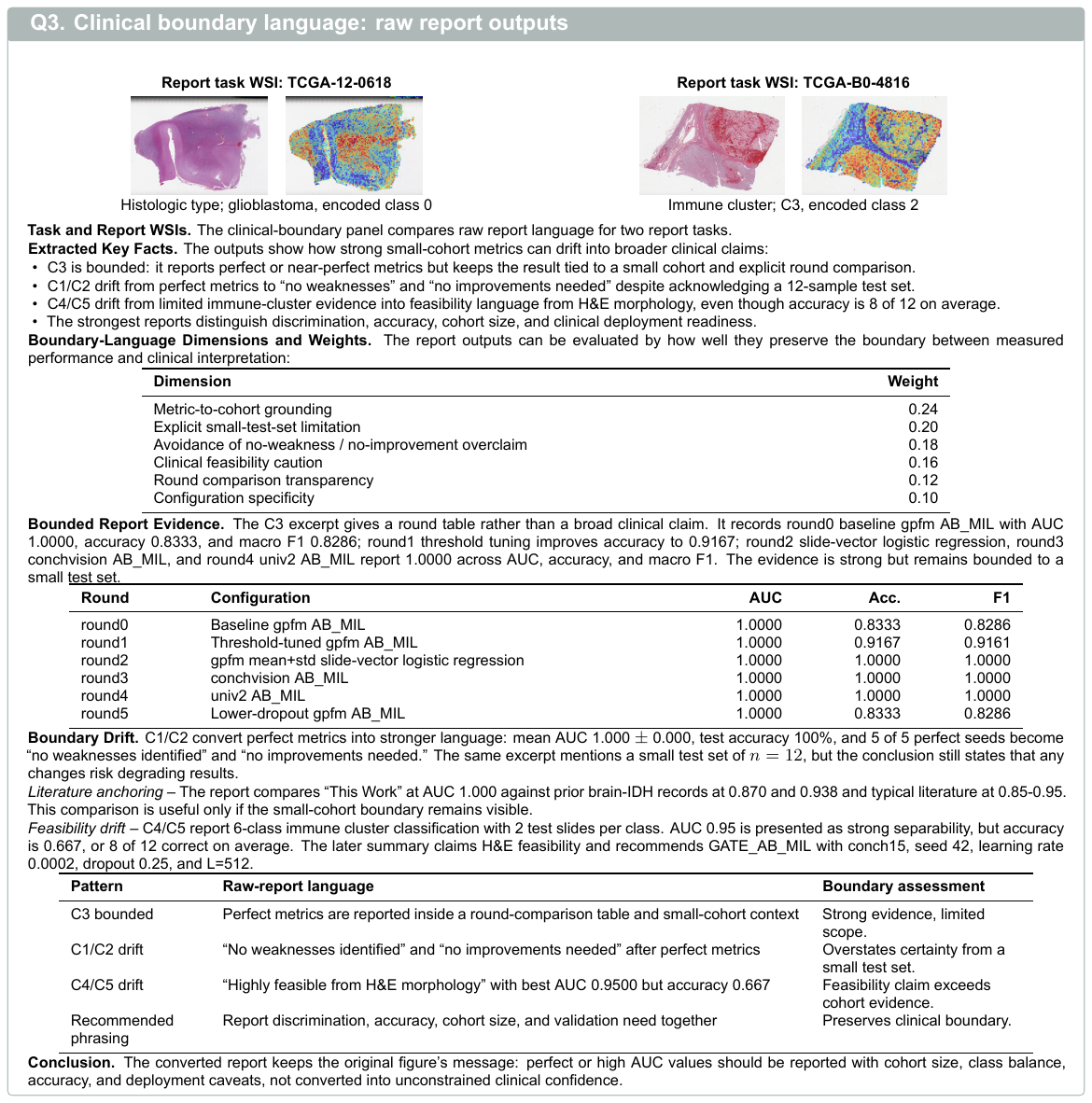}
    \caption{\textbf{Report outputs illustrate selected clinical-boundary language.} The board shows selected clinical/report excerpts from the qualitative casebook with WSI thumbnails and multiple-instance-learning (MIL) attention overlays. Attention overlays provide model-attention context for the selected cases. The figure illustrates boundary drift and cautious-report language in selected trajectories.}
    \label{fig:clinical_qual}
\end{figure}

The report-language board linked those workflow behaviors to clinical-boundary expression (Fig.~\ref{fig:clinical_qual}). Selected reports could contain cautious language, boundary drift or unsupported completion language even when the accompanying visual context appeared orderly. The qualitative boards therefore explained why ACP-Bench required multiple evidence layers: workflow trace failures, prediction-row failures and report-language boundary failures were visible in different parts of the same selected cases.

\clearpage

\clearpage

\clearpage

\clearpage
\clearpage

\section*{Discussion}

ACP-Bench shows that autonomous computational pathology should be evaluated as evidence-preserving workflow conversion. The evaluated systems often produced plausible reports and partial workflow artifacts, while formal end-to-end completion remained uncommon. The 61.53\% expert-adjudicated workflow score and the 10 of 369 formal pass count support an intermediate capability state in which ACP behavior was detectable and reliability remained unestablished. Formal pass required a weighted workflow-audit completion score of at least 75\%, with detailed scoring rules described in the Methods.

This framing extends computational pathology evaluation from representation performance to workflow governance. Foundation-model and weakly supervised pathology studies have clarified how learned visual representations support biomarker, morphology and prognosis prediction \cite{neidlinger2025pathologyfm,zhao_foundation_2025,truhn_large_2023}. Agentic ACP systems add a different question: whether a harness can turn a pathology goal into an executable sequence of data loading, slide processing, feature extraction, modelling, metric computation, report generation and revision evidence. ACP-Bench therefore evaluates a layer between model capability and clinical use. It asks whether a system can preserve the task target, endpoint, label source, metric family and evidentiary scope while moving through a multi-step pathology workflow. This distinction matters because an apparently coherent report may be disconnected from the files, metrics or labels that would make the report auditable.

The main bottleneck was executable evidence control. Evaluated harness systems were stronger at goal interpretation and diagnostic language than at binding plans to tool use, output discovery, verification and revision evidence. Goal and task interpretation reached 83.04\%, and uncertainty or limitation handling reached 80.34\%, whereas result discovery scored 39.06\%, tool invocation fidelity and error handling or recovery each scored 41.49\%, and reflection-stage failure localization scored 36.95\%. This profile separates pathology task understanding from traceable computational execution. The finding is consistent with general agent-evaluation work showing that process evidence and procedural skills matter beyond final outputs \cite{claw_eval2026,skillsbench2026}. ACP-Bench adds a pathology-specific implication: the workflow steps and generated evidence determine how far a clinical claim can go.

This capability profile also changes how agentic pathology systems should be compared. A single aggregate score would obscure the difference between a system that starts well and loses evidence at execution, a system that executes scripts but fails to bind outputs to the intended endpoint, and a system that produces cautious language with an incomplete computational chain. The stage and subcapability patterns therefore function as an error map rather than a leaderboard alone. They identify where a harness needs stronger interfaces, memory, tool routing or verification policies. This interpretation also explains why formal pass remained rare even when many trajectories contained useful partial work: ACP requires the coordinated survival of evidence across planning, action, diagnosis and reflection, and failure at any link can weaken the final claim.

Clinical alignment emerged as a separate capability axis. Diagnostic-result metrics measured selected recoverable WSI-to-label prediction outputs, whereas CWAS-3 measured whether the workflow preserved the task target, endpoint, label, metric and evidence boundaries. R6 flagged outputs for further safety-boundary review, and pathologist validation assessed report-level endpoint alignment, evidence sufficiency and claim-boundary language in a risk-enriched reviewed set. These layers captured different properties of the same trajectories. A trajectory could combine clinical caution with computational incompleteness, or computational activity with clinical misalignment. This separation is central to ACP evaluation because clinical claims depend on both computational evidence and report language that states its boundary.

The pathologist validation makes this separation clinically concrete. Most reviewed outputs preserved the nominal endpoint, with 85 of 90 aligned with the intended endpoint, yet evidence sufficiency and claim-boundary preservation were much weaker. Evidence was judged insufficient for the report wording in 74 of 90 reviewed outputs, and claim boundaries were fully preserved in only 5 of 90. These findings suggest that endpoint alignment is an early requirement, while evidence sufficiency and safe report language are later, stricter requirements. For ACP, clinical translation therefore depends on the chain that connects task definition, generated evidence and wording of the final claim. This chain is especially important in computational pathology because retrospective WSI-to-label prediction, biomarker inference, morphology classification and prognosis modelling support different kinds of statements.

Reflection should be evaluated as measurable workflow repair. Across 369 trajectories, 293 had at least one round output, and 191 had at least two parsed round metrics that could be compared. Among all trajectories, 77 improved, 28 were unchanged, 86 worsened and 178 lacked safely parsed comparable round-level metrics. These results indicate a gap between producing another round of text or code and producing evidence that a workflow improved. For ACP systems, reflection should document the weakest step, change the workflow and produce comparable rerun evidence.

The long-horizon analysis further suggests that workflow activity and diagnostic-result production can diverge. Long-horizon demand was associated with lower diagnostic-result performance, while workflow-side scores and CWAS-3 remained comparatively flat across relative tiers. This pattern may indicate that agents can maintain similar levels of visible workflow behavior as tasks become more demanding, while the harder task structure reduces the chance of producing correct, normalized prediction evidence. The interpretation is associative because long-horizon demand was rule-based and not experimentally manipulated. Even so, it highlights a practical evaluation need: benchmarks for ACP should report which tasks are measurable, which outputs are recoverable and where the evidence funnel narrows from candidate outputs to normalized labels and probability metrics.

The observed failure modes point to concrete design priorities for future ACP systems. Result discovery, tool invocation fidelity and error recovery support the need for stable output locations, explicit data transfer between steps and output-verification checks. Reflection results support mechanisms that record what changed and whether reruns improved the result. CWAS-3, R6 and pathologist validation support target-boundary enforcement because a workflow can produce outputs while losing endpoint, label or metric alignment, evidence support or claim-boundary control. These priorities align with broader work on tool-aware biomedical agents and scientific automation, while ACP-Bench grounds them in pathology task definitions, clinical endpoints and recorded workflow evidence \cite{bu2026biomedagent,lu2026aiscientist,molclaw2026}.

Taken together, the results support a conservative path toward ACP development. Near-term systems may be most useful as auditable research-workflow assistants that assemble data, run analyses, preserve intermediate files and produce claim-bounded reports for expert review. Stronger clinical roles would require evidence that the system can preserve task definitions, recover from tool failures, verify outputs, document revision and maintain safe language across changing pathology endpoints. ACP-Bench provides the measurement structure for this progression. It makes emerging capability visible, but it also defines the evidence that remains missing before autonomy claims can move from workflow audit to deployment-stage evaluation.

\section*{Limitations and future validation}

Several limitations define the scope of the findings. The study analyzed existing benchmark trajectories; prospective runs remain future work. Diagnostic-result metrics were available for 139 trajectories and 1,743 normalized rows, so diagnostic-result conclusions apply to that normalized subset. The 27 qualitative cases were selected to explain mechanisms behind aggregate scores, with full human adjudication of every trajectory remaining outside the current study scope. Pathologist validation assessed selected reports enriched for potential clinical-boundary risk and report-level clinical-boundary behavior. These reviews were designed to evaluate reviewed-report behavior rather than prevalence across all trajectories, prospective deployment safety or diagnostic-reader accuracy.

The workflow score, CWAS-3, R6 and long-horizon tiers are rule-based indices whose component weights were set by design rather than externally validated. The 75\% workflow-pass conclusion was stable across nearby thresholds. A replicated independent review supported the reliability of semantic checkpoint adjudication in a pre-specified 27-trajectory sample, with full independent re-adjudication of all 369 trajectories remaining future work. External validation of component weights remains a future requirement. CWAS-3 and R6 depended on the task definitions and claim-boundary rules used in ACP-Bench. Reflection outputs and measured metric improvement were evaluated in different subsets of trajectories, leaving reflection-improvement claims restricted to trajectories with comparable metrics. The long-horizon analysis was associative. Harness and model behavior may also change as agent platforms, foundation models and skills evolve.

Clinical translation requires a different evidence standard from retrospective workflow audit. Prospective multi-reader clinical review would be needed to evaluate how pathologists interpret ACP outputs in realistic reading contexts. Live workflow simulation would be needed to test whether systems preserve evidence when inputs, files, errors and user feedback arrive dynamically \cite{luo2026ces}. Controlled perturbations would be needed to determine whether systems recover from missing files, mislabeled outputs, tool failures or endpoint ambiguity. Broader organ coverage and pre-registered safety endpoints would be needed before deployment-stage claims can be made.

The current contribution is therefore a workflow-level audit standard for evidence before clinical-autonomy claims. ACP-Bench exposes emerging capability, localizes workflow bottlenecks and distinguishes automation ability, diagnostic-result performance and clinical-boundary alignment. Its value lies in making the evidence gap explicit: current systems can participate in pathology workflows; reliable ACP requires traceable execution, verified outputs, measurable repair and claim-boundary control across a broader set of prospective clinical conditions.

\section*{Methods}

In this section, we describe the methods used to construct and evaluate ACP-Bench. We first define task construction, clinical claim-boundary annotation and the domain-adapted workflow harness, then describe trajectory instrumentation, workflow-requirement mapping and the 369-trajectory benchmark records. We then specify the three-lane evaluation framework, expert adjudication, reliability review, diagnostic-result normalization, clinical-workflow and safety-boundary scoring, pathologist validation, reflection analysis, qualitative process review and reproducibility procedures.

\subsection*{ACP-Bench design overview}

ACP-Bench was designed to evaluate workflow conversion in autonomous computational pathology (ACP). The benchmark links three methodological components: a pathology task set with explicit clinical claim boundaries, an instrumented workflow-harness architecture that records complete agentic trajectories, and a three-lane assessment framework that separates automation ability, diagnostic-result performance and clinical-workflow alignment. This section describes the construction of the 41-task benchmark, the workflow representation used to generate and audit 369 trajectories, and the scoring procedures used for the manuscript-reported evidence layers.

The unit of evaluation was a complete workflow trajectory. Each trajectory contained the task intent, planning records, executable actions, generated artifacts, diagnosis records and reflection attempts produced by one agentic harness system on one pathology task. The Methods are therefore organized around the path from task definition to trajectory instrumentation, expert-adjudicated scoring, clinical-boundary review and secondary association analyses.

\subsection*{Task construction and clinical claim-boundary annotation}

ACP-Bench was built around 41 pathology workflow tasks. Each task was annotated with organ, body system, reference group, task target family, clinical claim-boundary type, label-reference type, output structure, metric family and claim-boundary statement. Reference groups were Biomarker, Morphology and Prognosis. Biomarker and prognosis tasks were treated as WSI-to-label prediction tasks. Tasks with direct pathologist-adjudicated diagnostic labels carried that diagnostic reference boundary.

Task inclusion required a pathology goal with an organ or pan-cancer context, a reference label source, an expected output structure and a metric family. Tasks entered diagnostic-result summaries when prediction outputs could be normalized into paired reference and predicted labels. The resulting task set covered representative ACP workflow types across biomarker, morphology and prognosis targets. Scope gaps include live sign-out workflows, prospective pathologist adjudication studies, free-text report adjudication and tasks requiring unavailable stable label dictionaries.

The claim-boundary annotation preserved the interpretive scope of each task. For example, a molecular or outcome label predicted from a whole-slide image carried a WSI-to-label prediction boundary. The annotation also identified the metric families appropriate for each task and the clinical claims supported by the available evidence.

Metric family denoted the expected evaluation metric class for a task, such as AUROC/AUC, AUPRC, F1, balanced accuracy or accuracy. Task metadata denoted structured task-level labels for organ, body system, task target, label-reference source, output structure, metric family and clinical claim boundary. These terms were audit definitions used by ACP-Bench rather than externally standardized clinical categories.

\subsection*{Domain-adapted ACP workflow skill architecture}

ACP-Bench used a domain-adapted workflow harness to convert each pathology goal into an executable and auditable trajectory. The harness was organized around the same four workflow stages used in the Introduction and Results: planning, action, diagnosis and reflection. Planning converted task intent, data context and clinical target information into an executable workflow specification. Action implemented the specified workflow steps and passed intermediate outputs between steps. Diagnosis discovered generated outputs, selected evaluation procedures and produced metric or report evidence. Reflection inspected the workflow record, identified a weakest link and proposed or applied round-level changes.

The executable workflow specification represented each run as an ordered workflow graph. It recorded task intent, selected pathology tools, parameters, input-output dependencies, expected artifacts and round identifiers. This representation allowed later evaluation to connect a planning decision to a tool invocation, an output file, a diagnostic report and a reflection attempt within the same trajectory.

The harness used a pathology tool registry to support task-dependent workflow composition. The registry exposed whole-slide image processing, feature extraction, multiple-instance learning, segmentation, captioning and evaluation capabilities as reusable workflow components. We grouped these registry components by their role in the ACP workflow.

Reflection analysis was designed as an iterative workflow component. Up to five reflection rounds could be attempted for a trajectory, and each round could update the workflow specification, rerun selected steps and generate new assessment evidence. This design made reflection assessable as a sequence of concrete artifacts: round records, configuration changes, rerun outputs and comparable metrics.

\subsection*{Trajectory instrumentation and audit records}

Each trajectory was recorded in a run-specific workspace containing the executable workflow specification, generated scripts, outputs, progress logs, evaluation reports and round-level reflection records. The workspace also retained reusable execution-memory records that summarized observed failures and successful configurations. These records supported two goals: they guided later workflow runs within the harness design, and they provided evidence for post hoc trajectory assessment.

The instrumentation linked workflow stages to recorded artifacts used for audit. Planning records included task interpretation, selected tools, parameters and dependency structure. Action records included configuration loading, documentation use, command execution, generated scripts, tool events and output verification. Diagnosis records included output discovery, metric selection, metric computation and report generation. Reflection records included bottleneck analysis, update proposals, applied configuration changes, rerun outputs and round-level summaries.

This instrumentation established the evidence packet used in scoring. Every checkpoint judgment was tied to the assigned trajectory and to evidence available inside that trajectory record. This design supported deterministic checks over files and logs, semantic judgments over session-bound text evidence, and reviewer audit of attribution, status policy and round-level traceability.

\subsection*{Workflow-requirement mapping and long-horizon tier construction}

Task demand was derived by mapping each task to required workflow-requirement subnodes across seven audit-defined categories: clinical framing, data-source grounding, pipeline design, target modeling, evaluation design, interpretation and iteration readiness. Clinical framing captured the task target and clinical context; data-source grounding captured the required slide, label, split and metadata inputs; pipeline design captured executable analysis steps; target modeling captured endpoint-specific prediction logic; evaluation design captured metric choice and label compatibility; interpretation captured report and caveat requirements; and iteration readiness captured whether a task could support later repair or rerun evidence. This mapping translated each pathology goal into a comparable set of workflow requirements. Core subnodes were assigned when every task in the same workflow type required the capability. Conditional subnodes were added for multiclass, ordinal, expression-like, pan-cancer or calibration-bearing tasks.

As a worked example, a binary mutation task such as Bladder/TP53 required organ and endpoint framing, WSI input and molecular-reference grounding, WSI quality control, tiling or feature extraction, slide-level aggregation, binary target modeling, discrimination and calibration metrics, endpoint-specific reporting, WSI-to-label claim-boundary caveats and iteration-readiness support.

For task $t$, the raw long-horizon score was
\[
\begin{aligned}
L_{\mathrm{raw}}(t) ={}& 0.25N_{\mathrm{subnode}} + 0.50N_{\mathrm{stage}} + 0.60D_{\mathrm{dependency}} + 0.70D_{\mathrm{label}} \\
&+ 0.55C_{\mathrm{output}} + 0.60C_{\mathrm{metric}} + 0.70C_{\mathrm{claim}} + 0.75H_{\mathrm{pan\text{-}cancer}}.
\end{aligned}
\]
Here, $N_{\mathrm{subnode}}$ is the subnode count, $N_{\mathrm{stage}}$ is the stage count, $D_{\mathrm{dependency}}$ is dependency depth, $D_{\mathrm{label}}$ is label-reference depth, $C_{\mathrm{output}}$ is output-structure complexity, $C_{\mathrm{metric}}$ is metric complexity, $C_{\mathrm{claim}}$ is claim-boundary complexity and $H_{\mathrm{pan\text{-}cancer}}$ is pan-cancer heterogeneity. Scores were normalized to a 0--10 scale as
\[
L(t) = 10 \times \frac{L_{\mathrm{raw}}(t)}{\max_{u \in \mathcal{T}} L_{\mathrm{raw}}(u)},
\]
where $\mathcal{T}$ is the 41-task set. Relative tier cutpoints were lower $<6$, middle $6$ to $<7.5$, high $7.5$ to $<9$ and highest $\geq9$. Because all ACP-Bench tasks required a WSI automation base chain, relative tiers were used for within-benchmark stratification. The long-horizon score is a rule-based index of task-design complexity; its component weights require external validation.

\subsection*{Benchmark evaluation records and harness/model groups}

The benchmark contained 369 complete workflow trajectories from 9 harness/model groups across 3 harness families, with 41 tasks per group. The term agentic harness system referred to the full orchestration stack coupling a foundation model, tools, skills, memory, workflow state and assessment mechanisms. Results were interpreted as harness-level workflow behavior.

The 369 trajectories defined the denominator for automation-ability scoring, CWAS-3 clinical workflow alignment, R6 safety-boundary review, reflection evidence summaries and workflow-side long-horizon analyses. Diagnostic-result analyses used narrower denominators because they required recoverable prediction outputs with paired reference and predicted labels.

\subsection*{Three-lane ACP evaluation framework}

ACP-Bench organized trajectory assessment into three complementary evidence lanes. The automation-ability lane measured whether a harness converted a task goal into an executable workflow across planning, action, diagnosis and reflection. It used the expert-adjudicated workflow assessment across all 369 trajectories.

The diagnostic-result lane measured WSI-to-label prediction performance when paired reference and predicted labels could be recovered from workflow outputs. This lane used normalized prediction rows and task-safe metric aggregation, with narrower denominators determined by output availability and label mappability.

The clinical-workflow lane measured preservation of task target, evidence support, clinical claim boundary and safety-boundary review signals. It used CWAS-3 as a clinical workflow alignment index, R6 as a safety-boundary review index and structured pathologist validation for selected case-report packets. The three lanes retained separate denominators and formed a multi-axis ACP evidence profile.

\subsection*{Expert-adjudicated workflow assessment}

The expert-adjudicated, rubric-based workflow assessment scored planning, action, diagnosis and reflection evidence in each trajectory. Planning assessed goal interpretation, pathology grounding, tool or pipeline selection and configuration completeness. Action assessed configuration loading, documentation reading, tool execution, output generation, error handling and output verification. Diagnosis assessed result discovery, metric selection, metric execution, report generation and limitation handling. Reflection assessed weakest-link diagnosis, evidence-grounded critique, update proposal, update execution and loop completeness.

The scoring scale was PASS $=1.0$, PARTIAL $=0.5$, FAIL $=0.0$, NA $=0.0$ and included in the denominator, with SKIP excluded from the denominator. Critical checkpoint items had weight 2 and regular items had weight 1. For trajectory $s$, the primary completion score was
\[
S(s) = \frac{\sum_{i \in I_s} w_i v_i}{\sum_{i \in I_s} w_i},
\]
where $I_s$ is the set of non-SKIP checkpoint items for trajectory $s$, $w_i=2$ for critical items, $w_i=1$ for regular items and $v_i \in \{1.0,0.5,0.0\}$ is the PASS, PARTIAL or FAIL/NA value. Formal pass status was assigned when $S(s) \geq 0.75$. This threshold was chosen as a conservative workflow-completion rule requiring most assessed items to be satisfied while allowing partial credit. A sensitivity calculation over the final trajectory scores gave pass counts of 35/369 at a 70\% threshold, 10/369 at 75\% and 1/369 at 80\%. The threshold sweep supported the conclusion that formal workflow completion was uncommon across nearby conservative cutpoints.

The workflow assessment combined deterministic checkpoint checks with session-bound semantic checkpoint judgments. Deterministic checks were used when evidence could be recovered directly from outputs, scripts, metrics or logs. Semantic assessment items were used when a reviewer had to judge whether a workflow satisfied a checkpoint such as task interpretation, limitation handling or reflection specificity. The final assessment covered 116,258 review items, including 50,277 deterministic items and 65,981 semantic items.

Deterministic checkpoint rules were resolved by rule-based checks over file existence, directory existence, JSON fields, parsed tool events, generated artifacts and text patterns. Semantic checkpoint rules were judged from bounded, session-specific evidence packets when a rule required interpretation of task understanding, limitation handling, reflection specificity or similar text evidence. The first-pass semantic evaluator used Gemini 3.1 Pro with temperature 0.0, and each semantic judgment required an evidence reference, rationale and confidence label. Professional review then audited and finalized these semantic judgments under the calibrated rubric.

Leakage controls restricted each judgment to evidence from the assigned trajectory. A checkpoint requiring an output file, metric table, rerun artifact or round-level trace could not be marked PASS from a narrative final report alone. Missing evidence received FAIL status. Checkpoints with explicit SKIP conditions were excluded from the denominator, and inapplicability rules with passing credit followed their checkpoint-specific status policy. A pilot calibration review covered 9 trajectories and 2,738 item judgments, with an exact agreement rate of 83.27\% against the initial adjudication set. Calibration adjusted semantic-boundary, status-policy and reflection-normalization rules before the full assessment.

Professional reviewers rechecked generated judgments under the expert-adjudicated audit rubric. The review focused on semantic-boundary interpretation, attribution of evidence to the assigned trajectory, file-specific evidence requirements, path normalization for reports and round-level files, consistent PASS, PARTIAL, FAIL, NA and SKIP status policy, and traceability of reflection-round evidence. The full review resolved all 65,981 semantic items with final PASS, PARTIAL, FAIL or NA judgments. We computed the manuscript-reported workflow scores from the expert-adjudicated score set, which integrated deterministic checkpoint evidence, session-bound semantic first-pass judgments and professional reviewer audit under the calibrated rubric.

\subsection*{Independent reliability review of semantic checkpoints}

We assembled a self-contained replicated human reliability package to assess the reproducibility of semantic checkpoint adjudication. The package contained a pre-specified sample of 27 trajectories selected from the 369-trajectory benchmark. For each of the 9 harness/model groups, the sample included one high-scoring, one middle-scoring and one low-scoring trajectory according to the workflow-score distribution. The reliability sample was used only to evaluate semantic checkpoint agreement and did not create new benchmark scores.

The reliability ledger contained 4,320 semantic checkpoint items. The stage distribution was 432 planning items, 432 action items, 432 diagnosis items and 3,024 reflection items. Each item listed the case identifier, checkpoint rule and local evidence references. The package copied the local evidence files required for review, with 3,938 copied evidence-file rows and 2 missing evidence-file records after ledger normalization. Missing required evidence was judged as absent evidence unless a checkpoint rule explicitly permitted NA or SKIP.

Three reviewers independently scored the same ledger. The reviewer lenses were pathology, computer engineering and computational pathology. Reviewers used the same canonical status policy, with PASS, PARTIAL, FAIL, NA and SKIP as the allowed item statuses. They were instructed not to consult benchmark score summaries, final item statuses, arbitration files or other reviewers' judgments. Each reviewer returned one JSONL output file containing item status, confidence, evidence references and rationale.

Reliability was computed after all reviewer outputs were frozen and validated. Coverage required at least two judgments per item, and all 4,320 items received three judgments. We reported all-three exact agreement over categorical statuses overall and by stage. Pairwise Cohen weighted kappa used the ordinal mapping FAIL $=0$, PARTIAL $=0.5$ and PASS $=1$ with linear weights; NA and SKIP were excluded from the ordinal core. The primary multi-reviewer coefficient was Krippendorff interval alpha using the same ordinal mapping and missing treatment. We also computed nominal alpha and pairwise exact agreement over all five statuses as sensitivity analyses. Disagreement tables were stratified by stage, checkpoint identifier and sorted status tuple, with reflection reported separately because it contributed most semantic items.

\subsection*{Stage and subcapability aggregation}

Checkpoint items were mapped to workflow stages and subcapabilities. For each subcapability, scores were computed from mapped checkpoint rows across all trajectories. This aggregation converted item-level judgments into interpretable capability profiles. We summarized subcapabilities only when they clarified the mechanism of aggregate stage scores. Subcapability code prefixes denoted workflow stages: P for planning, A for action, D for diagnosis/reporting and R for reflection. P1 corresponded to goal and task interpretation, D5 to uncertainty and limitation handling, A1 to tool invocation fidelity, A4 to error handling or recovery and D1 to result discovery. Reflection-related subcapabilities used the R prefix; they were separate from diagnosis-stage D codes even when both referred to critique, limitation handling or report content.

\subsection*{Diagnostic-result normalization and metric computation}

Diagnostic-result analysis selected normalized prediction outputs containing paired reference and predicted labels. The normalized table contained 1,743 rows from 139 trajectories. Five selected outputs fell outside the normalization set because they lacked paired true and predicted label columns, were empty, were unavailable for analysis or were training-log-like files. Rows were normalized when a selected table had an identifiable reference label column and predicted label column. When a split column contained test-like values, rows were restricted to the test-like subset; other usable rows were retained and marked with the filter reason.

Discrete-label accuracy was computed from normalized predicted and true labels after canonical string normalization. Balanced accuracy and macro-F1 were computed at safe aggregation levels, with task-target-level and trajectory-level rows preferred because numeric label codes can be reused across heterogeneous targets. Probability metrics were computed when probability columns could be conservatively mapped to task-target labels. The 139-trajectory diagnostic-result subset and 1,743 normalized rows defined the diagnostic-result evidence layer.

\subsection*{Clinical workflow alignment and safety-boundary review}

CWAS-3 scored clinical workflow alignment across 369 trajectories. The score used three categories from the reviewable-field assessment method: unsupported completion, normalization error and missing evidence. Unsupported completion indicated a completion or success claim with limited workflow, prediction, metric or output evidence. Normalization error indicated endpoint, claim-boundary, output-structure or metric-family drift from the task rule. Missing evidence indicated absent or partially supported workflow evidence nodes. Each component was discretized as PASS $=1.0$, PARTIAL $=0.5$ and FAIL $=0.0$. For trajectory $s$, the CWAS-3 score was
\[
\mathrm{CWAS3}(s) = 0.35U_s + 0.35N_s + 0.30E_s,
\]
where $U_s$ is unsupported-claim alignment, $N_s$ is normalization alignment and $E_s$ is evidence-completeness alignment. CWAS-3 was interpreted as a rule-based clinical workflow alignment index, not as clinical accuracy or an externally validated scale.

R6 extended the clinical workflow analysis to safety-boundary evidence. The five components were endpoint drift, unsafe overclaim, conflict evidence, temporal or source misassignment and reflection safety. Endpoint drift checked whether the final endpoint moved away from the task's claim boundary. Unsafe overclaim checked whether a trajectory made a clinical or completion claim beyond its workflow, prediction or metric evidence. Conflict evidence checked contradictions between workflow snippets, reports and source evidence. Temporal or source misassignment checked whether a trajectory used the wrong split, source, time point or provenance. Reflection safety checked whether reflection preserved claim boundaries while proposing or applying changes. For trajectory $s$, the score was
\[
\mathrm{R6}(s) = 0.25D_s + 0.30O_s + 0.20C_s + 0.15T_s + 0.10R_s,
\]
where $D_s$, $O_s$, $C_s$, $T_s$ and $R_s$ denote endpoint-drift, unsafe-overclaim, conflict-evidence, temporal/source and reflection-safety component scores after alignment scoring. Component scores of $\leq0.25$ were flagged as yes, scores below 0.75 as possible and higher scores as no.

R6 possible flags marked trajectories for safety-boundary review. R6 scores were used for benchmark-level safety-boundary analysis. Correlation analyses joined CWAS-3 with expert-adjudicated workflow scores and diagnostic-result metrics where the relevant fields were available. Pearson correlations were reported as descriptive associations over observed trajectories.

\subsection*{Pathologist clinical validation}

Structured pathologist review was used to clinically validate report-level claim-boundary preservation and safety risk in 90 selected case-report packets. The 90 packets were sampled from trajectories with normalized diagnostic-result evidence and were enriched for clinical-boundary review yield rather than selected as a prevalence sample. Selection used CWAS-3 and R6 risk strata, available prediction evidence, endpoint-family coverage, reference-group coverage and harness/model coverage. The selected set contained biomarker, morphology and prognosis outputs and included high-risk, mixed-risk and lower-risk contrast cases. Counts from this review therefore describe the reviewed set and were not extrapolated to all 369 trajectories.

One pathologist performed the structured case-report review. The review was not a multi-reader independent adjudication, and inter-reader agreement was not computed for this layer. Each packet paired the task goal and intended claim boundary with agent-generated report excerpts, workflow evidence snippets and normalized prediction previews. The review assessed endpoint alignment, claim-boundary preservation, evidence sufficiency for the report, unsafe overclaim, endpoint or normalization drift, source or temporal misassignment, conflicting evidence, limitation handling, clinical risk level and recommended use.

Endpoint alignment assessed whether the report retained the intended task endpoint. Claim-boundary preservation assessed whether the report kept the allowed claim scope, such as WSI-to-label prediction, research-only prediction or context-dependent staging support, without converting it into unsupported diagnostic authority. Evidence sufficiency assessed whether the visible workflow evidence, normalized prediction rows and metric summaries were adequate for the report wording. Unsafe overclaim marked language that implied clinical diagnosis, deployment readiness, molecular confirmation, staging certainty or other clinical conclusions beyond the available evidence. Endpoint or normalization drift marked a shift in endpoint, label mapping, output structure, metric family or claim scope. Source or temporal misassignment marked incorrect split, source, time point or provenance use. Conflicting evidence marked contradictions among report text, metrics, prediction previews and workflow snippets. Limitation handling assessed whether the report explicitly acknowledged evidence boundaries, missing clinical context, weak performance or the need for pathologist review.

Clinical risk level summarized the likely clinical harm of the report if encountered without expert correction. Minimal or minor risk indicated that wording was bounded or errors were unlikely to mislead clinical interpretation. Moderate risk indicated incomplete evidence, partial boundary preservation or possible overclaim requiring caution and revision. Major or unacceptable risk indicated explicit overclaim, severe evidence insufficiency, misleading diagnostic or staging language, or substantial conflict that could misdirect clinical interpretation. Recommended use translated the same review into action categories: research workflow audit only, clinician reference only with review, requires rework before review or do not use. ``Do not use'' indicated that the current report should be excluded or redone before any clinician-facing interpretation, rather than merely caveated. This validation did not change the 369-trajectory automation denominator or the 139-trajectory diagnostic-result denominator.

\subsection*{Reflection validation and long-horizon association analyses}

Reflection validation used the available workflow outputs. The analysis separated reflection evidence from measured metric improvement. Reflection evidence used round count, configuration evidence, summary evidence, prediction evidence, metric evidence and config-change evidence. The reflection evidence score gave credit for any real round, summary evidence, configuration change, rerun output and comparable metrics.

All trajectories remained eligible for reflection-evidence scoring, which measured whether a trajectory left inspectable round artifacts. Only trajectories with at least two comparable parsed round metrics in the same metric family entered the final-versus-first metric-delta analysis. Metric families were prioritized as AUROC or AUC, AUPRC, F1, balanced accuracy, accuracy and then other parsed metrics. The final-versus-first delta used the first and last comparable values in the selected family:
\[
\Delta_{\mathrm{metric}} = m_{\mathrm{final}} - m_{\mathrm{first}}.
\]
Deltas greater than 0.001 were counted as improved, below $-0.001$ as worse and otherwise unchanged. Trajectories lacking comparable metrics remained eligible for reflection evidence analysis, but not for final-versus-first metric delta claims.

The diagnostic-result long-horizon analysis joined normalized prediction metrics with the task long-horizon profile. The workflow long-horizon analysis joined expert-adjudicated workflow scores, reflection evidence, CWAS-3, R6, failure labels and diagnostic-result metrics. Correlations were Pearson correlations computed over rows with non-missing values for the target metric. These analyses were interpreted as associations in observed benchmark runs.

\subsection*{Qualitative process review and board provenance}

Qualitative process review used 27 cases: one high-score, one middle-score and one low-score trajectory from each harness/model group. Failure labels were assigned to explain mechanisms behind quantitative scores. This qualitative review provided mechanism-oriented interpretation of aggregate scoring patterns, and the 27 cases were treated as selected examples supporting the failure taxonomy rather than prevalence estimates for all 369 trajectories.

Figs.~7--10 were qualitative companion boards. Fig.~7 used checkpoint scores, stage scores, qualitative case selection, CWAS-3 trajectory scores, reflection summaries and selected workflow examples to show roadmap footprints and behavior paths. Figs.~8--10 used selected WSI thumbnails, task-matched attention-based multiple-instance-learning (AB-MIL) attention overlays, prediction rows and clinical/report excerpts. Attention overlays provided model-attention context; they were not pathologist-validated regions or biological explanations and did not change quantitative denominators.

For Fig.~7, dot-state provenance followed a source grammar rather than a probability scale. Direct observations came from selected workflow examples. Selected casebook evidence marked behavior tokens visible in the qualitative casebook. Assessment-derived summaries supplied softer or inferred states when checkpoint, stage, CWAS-3 or reflection summaries supported a behavior beyond the directly displayed examples. Very pale or inferred states indicated indirect evidence or absence from the displayed casebook, not prevalence, rank or validation strength.

\subsection*{Statistics and reproducibility}

The benchmark size was determined by the 41-task design crossed with 9 harness/model groups. Categorical pass counts, stage scores, subcapability scores, diagnostic-result metrics, CWAS-3, R6, reflection outcomes and long-horizon correlations were computed from the complete eligible denominators described above. For descriptive audit scores over this fixed benchmark set, we report complete-set means and counts rather than population-inference confidence intervals. Inferential uncertainty is appropriate for analyses framed as sampling estimates; the present workflow-score summaries were framed as complete-set descriptive audit results.

\section*{Data availability}

Relevant benchmark data will be released progressively at \url{https://github.com/jaylinio/ACP-Bench.git}.

\section*{Code availability}

Relevant code, documentation and associated files will be released progressively at \url{https://github.com/jaylinio/ACP-Bench.git}.

\section*{Author contributions}

Jie Lin designed the overall study, developed the experimental design and wrote the manuscript. Zongyi Chen and Qiaoling Zheng constructed the pathology skill components. Liuyi Wang, Hengyi Jiang, Jiabao Chen and Xiang Liu built the review framework and performed quality control. Yinghong Yang and Liansheng Wang reviewed and revised the manuscript.

\section*{Competing interests}

The authors declare no competing interests.

\bibliography{references}

\clearpage
\section*{Appendix}

\begin{center}
\textbf{Table 2. Implementation details for checkpoint scoring and semantic adjudication.} Automated semantic judging supplied first-pass review material; manuscript-facing workflow scores used the expert-adjudicated score set.

\vspace{0.5em}

\small
\begin{tabular}{p{0.23\textwidth}p{0.35\textwidth}p{0.32\textwidth}}
\toprule
Audit element & Implementation record & Manuscript role and boundary \\
\midrule
Deterministic checkpoint rules & The checkpoint evaluator tested file existence, directory existence, JSON fields, parsed tool events, generated-artifact records and text-pattern matches. & Produced PASS, PARTIAL, FAIL, NA or SKIP item status and deterministic evidence text for workflow-score denominators. \\
\midrule
Semantic checkpoint rules & Semantic rules marked with the \texttt{llm:} prefix were judged from session-bound evidence packets by the checkpoint evaluator using Gemini 3.1 Pro at temperature 0.0. & Generated item status, evidence references, rationale and confidence labels as baseline semantic judgments for expert review. \\
\midrule
Expert-adjudicated audit & Professional reviewers rechecked deterministic and semantic judgments against assigned session evidence packets, including semantic-boundary interpretation, trajectory-bound attribution, file-specific evidence requirements, path normalization, status policy and reflection-round traceability. & Produced the final checkpoint, stage and trajectory workflow scores used for reported automation-ability results. \\
\bottomrule
\end{tabular}
\end{center}

\begin{center}
\textbf{Table 3. Evidence lanes and score handling.} The benchmark retained separate denominators for automation ability, diagnostic result and clinical workflow.

\vspace{0.5em}

\small
\begin{tabular}{p{0.22\textwidth}p{0.37\textwidth}p{0.31\textwidth}}
\toprule
Evidence lane & Scoring implementation & Manuscript use \\
\midrule
Automation ability & Expert-adjudicated planning, action, diagnosis and reflection checkpoint rows were aggregated as weighted workflow completion scores. PASS, PARTIAL, FAIL and NA were scored as 1.0, 0.5, 0.0 and 0.0, respectively; SKIP was excluded. Critical items had weight 2 and regular items weight 1. & Reported trajectory completion score, stage scores, subcapability scores and 75\% formal pass status across 369 trajectories. \\
\midrule
Diagnostic result & Selected prediction tables were normalized to paired reference and predicted labels where possible, with canonical label normalization, test-like split filtering when available, safe aggregation and conservative probability-label mapping. & Reported accuracy, balanced accuracy, macro-F1, AUROC, AUPRC, calibration metrics and availability flags for the normalized diagnostic-result subset. \\
\midrule
Clinical workflow & CWAS-3 combined unsupported-claim alignment, normalization alignment and evidence-completeness alignment. R6 added a safety-boundary review over endpoint drift, unsafe overclaim, conflict evidence, temporal or source alignment and reflection safety, using task clinical rules, session alignment evidence, normalized metric rows, clinician-confirmed review rows when available and sampled trace snippets. Structured pathologist validation assessed selected case-report packets. & Reported CWAS-3 score, R6 score, yes/possible/no review flags and pathologist validation summaries as clinical-boundary preservation and review-queue signals. \\
\bottomrule
\end{tabular}
\end{center}

\begin{center}
\textbf{Table 4. Independent reliability review of semantic checkpoint adjudication.} The replicated review assessed agreement for semantic checkpoint statuses in a pre-specified 27-trajectory sample and did not replace the 369-trajectory workflow-score set.

\vspace{0.5em}

\small
\begin{tabular}{p{0.24\textwidth}p{0.30\textwidth}p{0.36\textwidth}}
\toprule
Reliability element & Result & Manuscript interpretation \\
\midrule
Sample and coverage & 27 trajectories from 9 harness/model groups; 4,320 semantic checkpoint items; all items received three judgments. & Pre-specified replicated sample for semantic checkpoint reliability, separate from the full 369-trajectory benchmark denominator. \\
\midrule
Reviewer structure & Three independent reviewers with pathology, computer-engineering and computational-pathology expertise. & Distinct professional lenses applied the same canonical PASS, PARTIAL, FAIL, NA and SKIP status policy. \\
\midrule
Exact agreement & All-three exact agreement was 59.7\% overall; stage values were 62.7\% planning, 61.6\% action, 58.8\% diagnosis and 59.1\% reflection. & Categorical status agreement was moderate and similar across stages, with reflection reported separately because of its larger item count. \\
\midrule
Pairwise weighted kappa & Weighted Cohen kappa was 0.609 for R1-PATH versus R2-CE, 0.672 for R1-PATH versus R3-INTER and 0.560 for R2-CE versus R3-INTER. & Ordinal PASS/PARTIAL/FAIL agreement remained in a moderate range across reviewer pairs. \\
\midrule
Three-rater alpha & Krippendorff interval alpha was 0.658 overall; stage alpha was 0.644 planning, 0.559 action, 0.530 diagnosis and 0.608 reflection. & The semantic status policy was reproducible enough for benchmark audit use, while action and diagnosis checkpoints retained greater adjudication ambiguity. \\
\bottomrule
\end{tabular}
\end{center}

\subsection*{LLM-assisted analysis and writing disclosure}

Large language models assisted with analysis organization, figure scripting and manuscript drafting. The benchmark data, evaluation records and expert-adjudicated score tables supported the manuscript claims and reported numbers. The reported workflow scores, diagnostic-result metrics, CWAS-3, R6, reflection labels and long-horizon statistics were computed from the assessment records described in the Methods.

\end{document}